\documentclass{article}
\usepackage{iclr2027_conference,times}

\usepackage{amsmath,amsfonts,bm}

\def\eqref#1{equation~\ref{#1}}

\def\1{\bm{1}}

\DeclareMathAlphabet{\mathsfit}{\encodingdefault}{\sfdefault}{m}{sl}
\SetMathAlphabet{\mathsfit}{bold}{\encodingdefault}{\sfdefault}{bx}{n}

\usepackage{hyperref}
\usepackage{url}

\usepackage[table]{xcolor}
\usepackage{booktabs}
\usepackage{multirow}
\usepackage{graphicx}
\usepackage{wrapfig}
\usepackage{caption}
\usepackage{titletoc}

\newcommand{\up}[1]{\textcolor{green!60!black}{\scriptsize$\uparrow$#1}}

\title{MLLMs Know When Before Speaking: Revealing and Recovering Temporal Grounding via Attention Cues}

\makeatletter
\newcommand{\symfootnote}[1]{\begingroup\renewcommand\thefootnote{}\renewcommand\@makefnmark{}\footnotetext{#1}\endgroup}
\makeatother
\author{%
\bfseries Dazhao Du\textsuperscript{1,3,$\ast$},\enspace Liao Duan\textsuperscript{1,2,$\ast$},\enspace Jian Liu\textsuperscript{1},\enspace Tao Han\textsuperscript{1},\enspace Yujia Zhang\textsuperscript{3},\enspace Eric Liu\textsuperscript{3},\enspace Xi Chen\textsuperscript{3},\\[2pt]
\bfseries Song Guo\textsuperscript{1,$\dagger$}\\[5pt]
{\normalfont\normalsize\textsuperscript{1}Hong Kong University of Science and Technology\quad\textsuperscript{2}Xi'an Jiaotong University\quad\textsuperscript{3}Tencent}%
}

\iclrfinalcopy
\begin{document}
\maketitle
\lhead{Preprint}
\symfootnote{$^{\ast}$Equal contribution. Part of this work was done while Dazhao was an intern at Tencent and Liao was a RA at HKUST.}
\symfootnote{$^{\dagger}$Corresponding author.}

\maketitle

\begin{abstract}
Video temporal grounding (VTG), which localizes the start and end times of a queried event in an untrimmed video, is a key test of whether multimodal large language models (MLLMs) understand not only what happens but also when it happens. Although modern MLLMs describe video content fluently, their timestamp predictions remain unreliable, while existing remedies either require costly post-training on temporal annotations or rely on coarse training-free heuristics. In this work, we probe the cross-modal attention of MLLMs and uncover a perception-generation gap. Our key finding is that MLLMs often know the target interval during prefill, but lose this signal when generating the final answer. In the prefill stage, a sparse set of attention heads, which we call \emph{Temporal Grounding Heads} (TG-Heads), concentrates query-to-video attention on the ground-truth interval. During autoregressive decoding, however, the answer tokens shift attention away from this interval toward visually salient but query-irrelevant segments. This observation motivates an inference-time read-then-regenerate framework. We first convert TG-Head prefill attention into a debiased frame-level relevance signal and extract the high-attention interval it highlights. We then re-invoke the MLLM with visual context restricted to this interval using video cropping. Without parameter updates or architectural changes, our framework consistently improves MiMo-VL, Qwen3-VL, and Molmo2 on three benchmarks, with gains of up to +3.5 mIoU. The project website can be found at \url{https://ddz16.github.io/mllmsknowwhen.github.io/}.
\end{abstract}

\section{Introduction}

Multimodal large language models (MLLMs) have become remarkably capable video understanders~\citep{videollama,li2024videochat,comanici2025gemini,bai2025qwen3}, yet a simple follow-up question often trips them up: \emph{when does it happen?} Video temporal grounding (VTG), which requires predicting the start and end times of a queried event in an untrimmed video, remains a persistent weakness of general-purpose MLLMs~\citep{wu2025survey}. Existing remedies expose an uncomfortable trade-off. Post-training methods improve VTG by fitting models on large temporal-annotation corpora~\citep{zhang2025timelens, wang2025time, zeng2024timesuite, chen2025datasets}, but require expensive annotations and specialized training pipelines. Training-free alternatives avoid this cost, but typically replace temporal localization with indirect surrogates such as segment-caption matching or numeric frame reading, sacrificing fine-grained boundary precision~\citep{zheng2024training, wu2025number, xu2025zero}. This raises the question that motivates our work: \emph{do MLLMs genuinely lack temporal grounding ability, or do they already possess it in a latent form that standard decoding fails to surface?}

\begin{figure}[t]
  \centering
  \includegraphics[width=\linewidth]{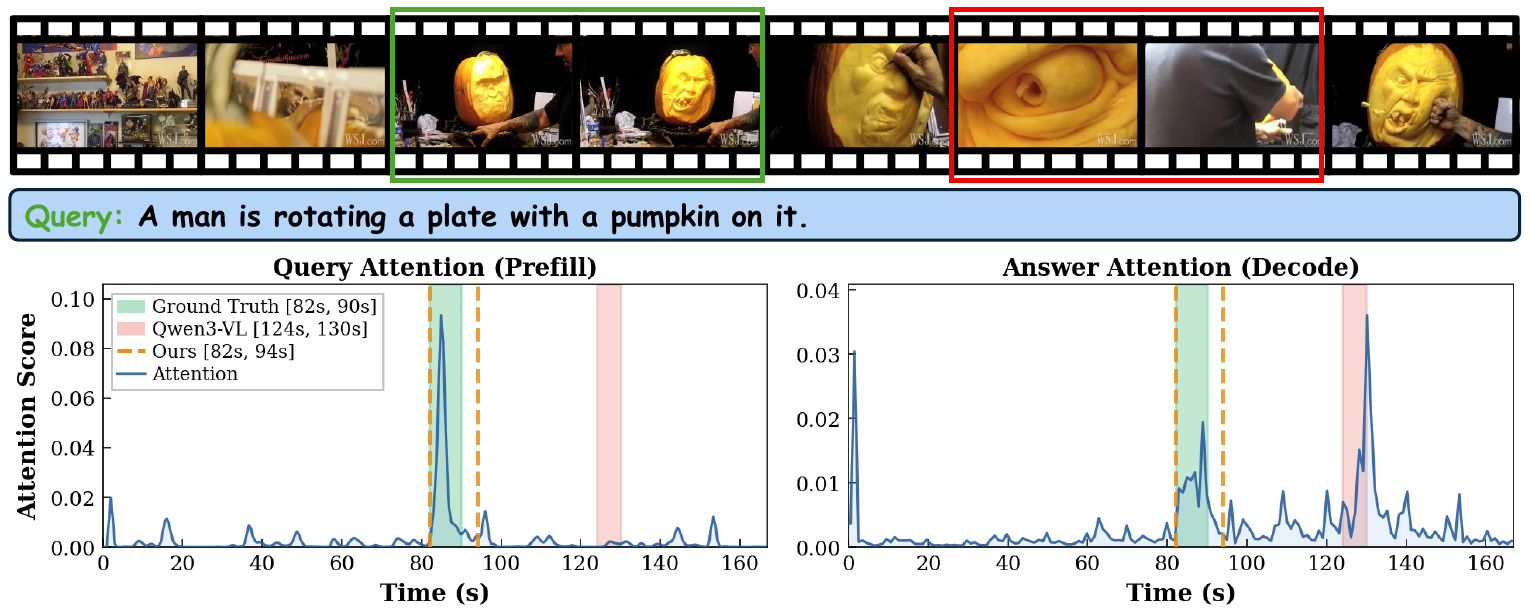}
  \caption{\textbf{MLLMs know \emph{when} during prefill but forget during decoding.} We visualize Qwen3-VL's cross-modal attention at inference. During prefill (left), attention from query tokens peaks at the ground-truth interval. During decoding (right), attention from the generated answer tokens drifts away to a visually salient but query-irrelevant segment, which aligns with the model's erroneous prediction. Our method leverages the clean prefill signal to correct the final output.}
  \label{fig:motivation}
\end{figure}

To test this possibility, we first ask where a latent temporal signal would reside inside the model. Recent studies suggest that attention heads in MLLMs often specialize into distinct functional roles~\citep{kang2025your, baek2025large, bi2025unveiling, jiang2025investigating}. We therefore conduct a systematic \emph{head knockout} study~\citep{zhang2025cross,gou2025empirical}: for each head, we suppress its query-to-video attention and measure the resulting drop in VTG accuracy. The result is sharply non-uniform. Out of hundreds of heads, only a small subset causes a substantial degradation when removed (Figure~\ref{fig:TG_heads}). We call them \emph{Temporal Grounding Heads} (TG-Heads). This intervention evidence suggests that MLLMs do contain an internal temporal localization mechanism, but that it is concentrated in a sparse set of heads rather than distributed uniformly across the network.

Having located TG-Heads, we next ask how their signal is used during inference. We compare their cross-modal attention at two stages: \emph{prefill}, where query tokens attend over the video, and \emph{autoregressive decoding}, where newly generated answer tokens attend back to the same visual context. This reveals a clear perception-generation gap (Figure~\ref{fig:motivation}). During prefill, TG-Head attention from query tokens concentrates on the ground-truth interval, suggesting that the relevant frames have already been identified. During decoding, however, attention from the numeric answer tokens shifts away from this interval toward visually salient but query-irrelevant segments, and the final timestamp follows this misdirected attention. More examples are provided in Appendix~\ref{sec:motivation_appendix}. This kind of disconnect between internal visual evidence and final output echoes recent observations in image VLMs~\citep{liu2025seeing, zhang2025mllms}, but here it appears along the temporal axis of video.

We hypothesize that this gap arises from an asymmetry between reading and speaking. During prefill, the full query provides a focused linguistic intent, allowing TG-Heads to bind discriminative words and actions to a small set of relevant frames. During decoding, the model instead generates timestamp tokens, whose numeric form provides little semantic guidance while still attending over hundreds of visual tokens. The localized signal established by TG-Heads is therefore diluted within a much larger visual context and can be overwhelmed by salient distractors. The model knows \emph{when}, but forgets at the moment it has to speak.

This diagnosis suggests a direct intervention: read out the clean prefill signal before it is lost, and let the model speak again after the distracting visual context has been removed. We instantiate this idea as an inference-time \emph{read-then-regenerate} framework. Stage~1 extracts and refines the TG-Head prefill attention to detect a high-attention temporal interval. Stage~2 re-invokes the MLLM with visual context restricted to the detected interval by cropping the video. A confidence gate skips this second stage when the original prediction is already confident enough or aligns with the attention evidence. Across MiMo-VL~\citep{coreteam2025mimovltechnicalreport}, Qwen3-VL~\citep{bai2025qwen3}, and Molmo2~\citep{clark2026molmo2} on three TimeLens-Bench datasets~\citep{zhang2025timelens}, our method consistently improves VTG performance by up to +3.5 mIoU, without parameter updates or architectural changes.

Our contributions are summarized as follows:
\begin{itemize}
    \item We identify \emph{Temporal Grounding Heads}, a sparse subset of attention heads whose query-to-video prefill attention provides intervention-backed evidence of latent temporal localization inside MLLMs.
    \item Through these heads, we reveal a perception-generation gap on VTG: the model knows the correct interval during prefill but is distracted by irrelevant frames during decoding.
    \item Motivated by this diagnosis, we propose an inference-time \emph{read-then-regenerate} framework that recovers this hidden signal and improves three MLLMs on three TimeLens-Bench datasets without parameter updates or architectural changes.
\end{itemize}

\section{Related Work}
\label{sec:related_work}

\paragraph{MLLMs for Temporal Grounding.}
Existing work on video temporal grounding with MLLMs can be broadly divided into post-training and training-free paradigms. \emph{Post-training} methods~\citep{chen2025datasets,cheng2025tempura} either introduce specialized architectural components for temporal encoding, such as the learnable timestamp modality in Tempo-R0~\citep{yue2025tempo} or the 1D temporal convolutions in TimeSuite~\citep{zeng2024timesuite}, or rely on large-scale supervised fine-tuning and reinforcement learning on VTG corpora~\citep{li2025tempsamp,wang2025time,wang2024grounded}. A recent representative is TimeLens~\citep{zhang2025timelens}, which post-trains MLLMs on a re-annotated 100K corpus with thinking-free RLVR. Although effective on VTG benchmarks, post-training incurs substantial annotation and compute costs and can degrade general video understanding. \emph{Training-free} methods~\citep{zheng2024training,xu2025zero} avoid these costs but often rely on indirect cues: ChatVTG~\citep{qu2024chatvtg} and VTG-GPT~\citep{xu2024vtg} reformulate VTG as text-level matching between query and segment captions, while NumPro~\citep{wu2025number} burns numeric watermarks onto frames and turns temporal prediction into OCR-style reading. These approaches convert continuous temporal localization into discrete or coarse-grained surrogates, which can limit boundary precision. In contrast, our method uses the model's own cross-modal attention as an internal temporal cue: after one-time TG-Head identification, it performs inference-time read-then-regenerate without parameter updates or architectural changes, yielding fine-grained localization across multiple general-purpose MLLM families.

\paragraph{Attention Analysis in MLLMs.}
A growing body of work studies the internal attention of MLLMs and finds that it often reflects the correct cross-modal grounding even when the final output is wrong~\citep{liu2025seeing, zhang2025mllms}. Further analyses reveal that this visual information is not uniformly distributed across the network, but is instead carried by a sparse set of functional heads that specialize in distinct semantic roles, such as text-image alignment~\citep{bi2025unveiling, jiang2025investigating,kim2025interpreting}, OCR-like reading~\citep{baek2025large}, and spatial grounding~\citep{kang2025your}. In the video domain, deep attention layers have been observed to encode query-conditioned keyframe priors~\citep{li2025less}, suggesting that MLLMs may already contain a latent temporal grounding signal. How to reliably extract and exploit this signal for precise boundary prediction, however, remains open. We address this gap by identifying \emph{Temporal Grounding Heads} via a systematic head knockout study~\citep{zhang2025cross,gou2025empirical} and translating their attention into explicit temporal intervals without any additional training.

\section{Temporal Grounding Head}
\label{sec:tghead}

Recent studies have shown that internal attention in MLLMs exhibits clear functional specialization, with different heads playing different roles~\citep{kang2025your, bi2025unveiling, jiang2025investigating, baek2025large}. Motivated by these observations, we ask whether temporal grounding is likewise supported by a specific subset of heads. To find out, we probe each head in the MLLM with \emph{attention knockout}: we block the cross-modal attention flow through a single head and measure how much the grounding performance drops. Heads whose removal causes a disproportionate degradation are flagged as the functional units responsible for temporal localization.

\begin{figure}[t]
  \centering
  \includegraphics[width=1.0\linewidth]{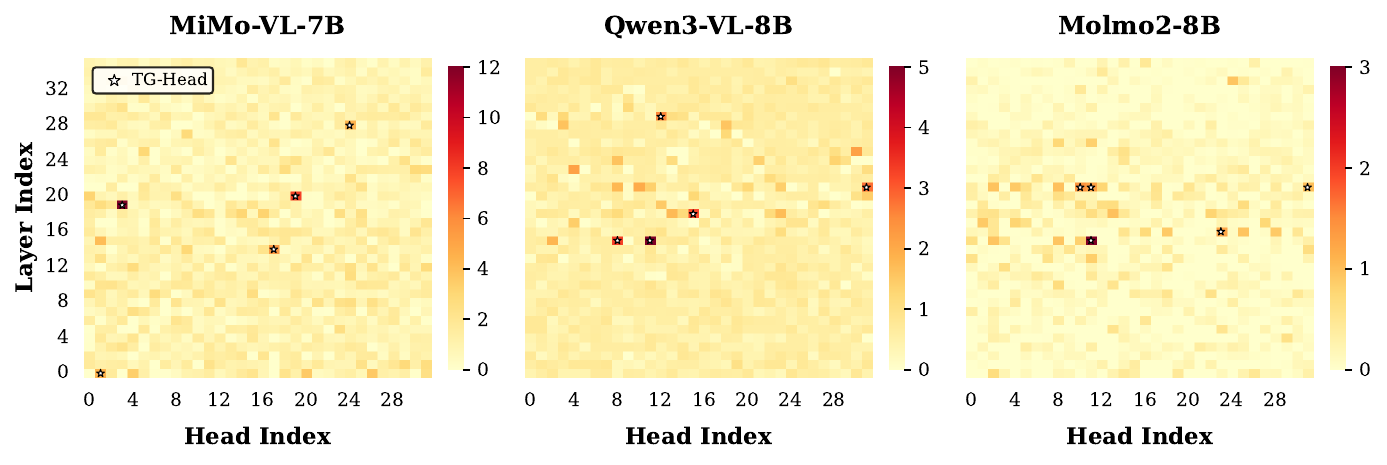}
  \caption{\textbf{Grounding Contribution Score (GCS) of each attention head} across three MLLMs. Each dot is one head, and the top-$K$ heads are marked with stars.}
  \label{fig:TG_heads}
\end{figure}

\paragraph{Attention knockout.}
Let the MLLM have $L$ layers and $H$ heads per layer. For head $h$ at layer $\ell$, the attention weights are computed as
\begin{equation}
\mathbf{A}^{\ell,h} \;=\; \operatorname{softmax}\!\left(\frac{\mathbf{Q}^{\ell,h}(\mathbf{K}^{\ell,h})^{\top}}{\sqrt{d_k}} + \mathbf{M}^{\ell,h}\right),
\label{eq:attn}
\end{equation}
where $\mathbf{Q}^{\ell,h}$ and $\mathbf{K}^{\ell,h}$ are the query and key matrices of head $(\ell,h)$, $d_k$ is the per-head feature dimension that normalizes the dot-product magnitude, and $\mathbf{M}^{\ell,h}$ is an additive mask with the same shape as $\mathbf{Q}^{\ell,h}(\mathbf{K}^{\ell,h})^{\top}$ that controls which attention edges are allowed. In a standard autoregressive transformer, $\mathbf{M}^{\ell,h}$ is the causal mask, i.e., $\mathbf{M}^{\ell,h}_{s,t}=0$ if $t\leq s$ and $-\infty$ otherwise, which prevents each position from attending to future positions.

Inspired by prior attention-knockout analyses of LLMs and MLLMs~\citep{zhang2025cross, gou2025empirical}, we selectively block the cross-modal attention edges from query (text) positions to video positions for a single head. Let $\mathcal{T}=[Q_s,Q_e)$ denote the set of query-token positions and $\mathcal{S}=[V_s,V_e)$ denote the set of video-token positions. To knock out head $(\ell,h)$, we modify its attention mask to:
\begin{equation}
\widetilde{\mathbf{M}}^{\ell,h}_{s,t} \;=\;
\begin{cases}
-\infty, & \text{if } t > s, \\
-\infty, & \text{if } s \in \mathcal{T} \ \text{and}\ t \in \mathcal{S}, \\
0, & \text{otherwise},
\end{cases}
\label{eq:knockout}
\end{equation}
while keeping all other heads unchanged. Recall that $s$ indexes the query (row) dimension and $t$ the key (column) dimension. The first case is the standard causal constraint. The second sets the attention from any query-text position $s\in\mathcal{T}$ to any video position $t\in\mathcal{S}$ to zero after softmax, surgically removing the cross-modal information flow through this head without disturbing intra-modal attention.

\paragraph{Measuring each head's contribution.}
Intuitively, if a head is truly important for temporal localization, then blocking its cross-modal attention edges should make the model's grounding predictions noticeably worse. Conversely, if a head is irrelevant (or even noisy) for this task, blocking it should have little effect, or may even slightly improve accuracy. We therefore quantify the contribution of each head by how much grounding performance drops when that single head is knocked out. Concretely, we evaluate the MLLM on a held-out VTG validation subset and measure the mean Intersection over Union (mIoU): let $\mathcal{M}_{\text{base}}$ be the mIoU of the intact model, and let $\mathcal{M}_{\ell,h}$ be the mIoU obtained after applying Eq.~(\ref{eq:knockout}) to head $(\ell,h)$. We define the \emph{Grounding Contribution Score} of head $(\ell,h)$ as the resulting mIoU drop:
\begin{equation}
\operatorname{GCS}_{\ell,h} \;=\; \mathcal{M}_{\text{base}} - \mathcal{M}_{\ell,h}.
\end{equation}
A large positive $\operatorname{GCS}_{\ell,h}$ means that the model heavily relies on head $(\ell,h)$ for temporal grounding, whereas a near-zero or negative value means the head carries little temporal grounding information.

\paragraph{TG-Head selection.}
Figure~\ref{fig:TG_heads} visualizes $\operatorname{GCS}_{\ell,h}$ across the full $L\times H$ grid for three MLLMs. In every model, the distribution is sharply heavy-tailed: the vast majority of heads yield near-zero or even slightly negative GCS, while only a small number of heads contribute substantially. The long-tailed pattern is also confirmed by the sorted pruning curves in Appendix~\ref{sec:mIOU_drop_appendix}. This confirms the existence of a highly specialized subset of heads dedicated to temporal grounding. We therefore simply select the top-$K$ heads with the highest GCS as our \emph{Temporal Grounding Heads} (TG-Heads). Importantly, this is a one-time model-level calibration using only a small labeled subset rather than model training requiring the full VTG training corpora (see Appendix~\ref{sec:impl_details_appendix} and Appendix~\ref{sec:calibration_stability}).

\section{Inference-Time Read-Then-Regenerate with Attention Cues}
\label{sec:method}

Given a video $\mathcal{V}$ of duration $D$ consisting of $T$ sampled frames $\{v_1,\dots,v_T\}$, and a natural language query $q=(q_1,\dots,q_M)$ of length $M$, our goal is to predict a temporal interval $[\hat{s},\hat{e}]\subset[0,D]$ that localizes the queried event. Let $\Delta t=D/T$ denote the per-frame duration. Let $\mathcal{H}=\{(\ell_1,h_1),\dots,(\ell_K,h_K)\}$ denote the set of $K$ TG-Heads identified in Section~\ref{sec:tghead}. Our framework proceeds in two stages.

\begin{figure}[t]
  \centering
  \includegraphics[width=1.0\linewidth]{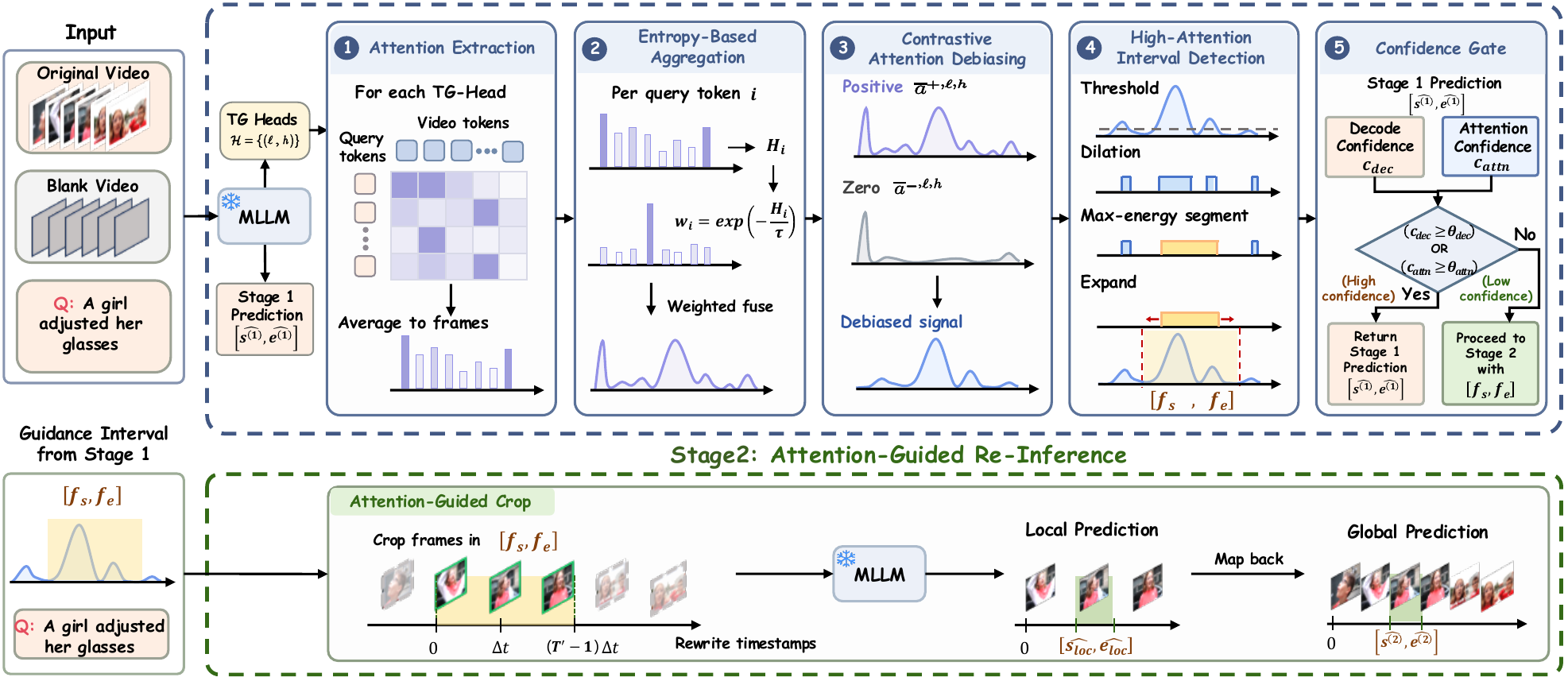}
  \caption{\textbf{Overview of our two-stage framework.} \textbf{Stage 1} runs the MLLM once to obtain a baseline prediction $[\hat{s}^{(1)},\hat{e}^{(1)}]$ and, in parallel, processes the TG-Head prefill attention through (1) extraction, (2) entropy-based aggregation, (3) contrastive debiasing against a blank-video reference, and (4) high-attention interval detection. (5) A confidence gate decides whether the baseline prediction is already reliable. \textbf{Stage 2} crops the video to the detected interval $[f_s,f_e]$, re-invokes the MLLM on the focused clip.}
  \label{fig:method}
\end{figure}

\subsection{Stage 1: Initial Inference}

\paragraph{Attention extraction.}
We perform a single forward pass of the MLLM on $(\mathcal{V}, q)$. This pass has two outputs. First, by letting the MLLM greedily decode an answer and parsing the emitted numeric tokens, we obtain a \emph{Stage 1 prediction} $[\hat{s}^{(1)}, \hat{e}^{(1)}]\subset[0, D]$. Second, during the prefill stage of this same pass, we read out the cross-modal attention weights from each TG-Head $(\ell,h)\in\mathcal{H}$. Following Eq.~(\ref{eq:attn}) with the standard causal mask, we denote the full attention weights at head $(\ell,h)$ by $\mathbf{A}^{\ell,h}$. Slicing the rows corresponding to the query-token positions $\mathcal{T}=[Q_s,Q_e)$ and the columns corresponding to the video-token positions $\mathcal{S}=[V_s,V_e)$ yields the query-to-video attention:
\begin{equation}
\mathbf{A}^{\ell,h}_{q\to v} \;=\; \mathbf{A}^{\ell,h}_{\mathcal{T},\mathcal{S}} \;\in\; \mathbb{R}^{M\times N_v},
\label{eq:qv_attn}
\end{equation}
where $N_v=|\mathcal{S}|$ is the total number of video tokens. Each row of $\mathbf{A}^{\ell,h}_{q\to v}$ is the attention distribution of one query token over all video tokens. 


\paragraph{Entropy-based attention aggregation.}
To obtain a single frame-level curve that measures the query-video relevance, we need to reduce $\mathbf{A}^{\ell,h}_{q\to v}$ along two axes. Along the video-token axis, we simply average the attention values of all visual tokens belonging to the same frame, yielding a query-token-to-frame distribution $\mathbf{a}_i^{\ell,h}\in\mathbb{R}^T$ for each query token $i$. Along the query-token axis, the $M$ distributions are however not equally informative: function words such as ``a'' or ``the'' spread attention almost uniformly over the video, while content words (nouns, verbs) concentrate attention on the semantically relevant frames (see Appendix~\ref{sec:entropy_appendix}). Uniformly averaging over all $M$ rows would therefore be dominated by the uninformative ones. We instead weight each row by the negative exponential of its entropy: we normalize $\mathbf{a}_i^{\ell,h}$ to a probability distribution, compute its Shannon entropy $H_i$, and assign the weight $w_i=\exp(-H_i/\tau)$ with temperature $\tau{=}0.5$. This up-weights concentrated (low-entropy) rows and down-weights diffuse (high-entropy) ones. The fused query-video relevance curve for head $(\ell,h)$ is:
\begin{equation}
\bar{\mathbf{a}}^{\ell,h}(j) \;=\; \frac{\sum_{i=1}^{M} w_i\,\mathbf{a}_i^{\ell,h}(j)}{\sum_{i=1}^{M} w_i}, \qquad j=1,\dots,T.
\label{eq:entropy_agg}
\end{equation}

\paragraph{Contrastive attention debiasing.}
Even after entropy-based aggregation, the curve $\bar{\mathbf{a}}^{\ell,h}$ still contains spurious peaks that are not driven by query-specific content. A representative example is the \emph{attention sink}~\citep{yu2024unveiling, kang2025see}: irrespective of the query, MLLMs consistently allocate a large fraction of attention to a few positional anchors such as the first visual tokens of the video, which can dominate the detected interval (see Appendix~\ref{sec:debiasing_appendix}). To remove such content-independent biases, we borrow the idea of contrastive decoding~\citep{li2023contrastive}: the bias part is isolated by running the same procedure on a reference input that carries no semantic content, and subtracted from the original signal. Concretely, we replace all pixel values of $\mathcal{V}$ with zeros to obtain a blank video $\mathcal{V}^{-}$, run a second forward pass on $(\mathcal{V}^{-}, q)$, and apply the same token-to-frame and entropy aggregation procedure, yielding a \emph{zero-video} curve $\bar{\mathbf{a}}^{-,\ell,h}$. We denote the curve obtained from the original video (Eq.~\ref{eq:entropy_agg}) as $\bar{\mathbf{a}}^{+,\ell,h}$. Both curves are Gaussian-smoothed and $L_1$-normalized. The per-head debiased signal is then the pointwise positive log-ratio:
\begin{equation}
\mathbf{s}^{\ell,h}(j) \;=\; \max\!\left(\bar{\mathbf{a}}^{+,\ell,h}(j)\,\log\frac{\bar{\mathbf{a}}^{+,\ell,h}(j)}{\bar{\mathbf{a}}^{-,\ell,h}(j)},\; 0\right),
\label{eq:debias_head}
\end{equation}
which retains only the frames where the positive video has higher attention than what the bias alone would predict. Finally, we fuse across the $K$ TG-Heads by a simple average:
\begin{equation}
\mathbf{s}(j) \;=\; \frac{1}{K}\sum_{(\ell,h)\in\mathcal{H}} \mathbf{s}^{\ell,h}(j).
\label{eq:debias_avg}
\end{equation}
The resulting $\mathbf{s}\in\mathbb{R}^T$ is the single debiased frame-level relevance signal used below.

\paragraph{High-attention interval detection.}
From $\mathbf{s}$ we extract a temporal interval that captures the query-relevant region. We adopt a simple threshold-then-dilate routine, analogous to how prior work extracts spatial bounding boxes from VLM attention maps via thresholding and connected-region analysis~\citep{kang2025your}, but adapted to the 1D temporal axis. The routine has four steps. First, we binarize the signal by keeping frames where $\mathbf{s}(j)\ge\rho\cdot\max_j\mathbf{s}(j)$. Second, we perform 1D morphological dilation with a window of $\lceil\delta/\Delta t\rceil$ frames, which sets a frame to active whenever any frame within the window is active, thereby bridging short gaps between nearby active regions that likely belong to the same event. Third, among the resulting connected segments we select the one with the largest cumulative energy $\sum_{j\in\text{seg}}\mathbf{s}(j)$. Finally, we expand the chosen segment by a fraction $r$ of its length on each side (clamped to $[1,T]$) to provide temporal context. The output is the frame interval $[f_s,f_e]\subset[1,T]$, corresponding to the time interval $[(f_s-1)\Delta t,\; f_e\,\Delta t]$.

\paragraph{Confidence gate.}
Finally, we estimate whether the Stage 1 prediction $[\hat{s}^{(1)},\hat{e}^{(1)}]$ already aligns with the attention signal, in which case Stage 2 is unnecessary. We compute two complementary confidence scores. The \emph{decode confidence} $c_{\text{dec}}$ is the geometric mean of the softmax probabilities of all numeric tokens emitted during Stage 1 greedy decoding: $c_{\text{dec}}=\exp\!\big(\tfrac{1}{|\mathcal{N}|}\sum_{i\in\mathcal{N}}\log p_i\big)$, where $\mathcal{N}$ is the set of numeric-token positions in the generated answer and $p_i$ is the softmax probability of the chosen token at position $i$. The \emph{attention confidence} $c_{\text{attn}}$ measures whether the predicted region already contains the attention peak: $c_{\text{attn}}=\max_{j\in[\hat{f}_s^{(1)},\hat{f}_e^{(1)}]}\mathbf{s}(j)\,/\,\max_{j}\mathbf{s}(j)$, where $[\hat{f}_s^{(1)},\hat{f}_e^{(1)}]$ is the frame interval corresponding to the Stage 1 prediction (in the rare case where $\max_j\mathbf{s}(j)=0$, we set $c_{\text{attn}}=0$). If either score exceeds its threshold, $c_{\text{dec}}\ge\theta_{\text{dec}}$ or $c_{\text{attn}}\ge\theta_{\text{attn}}$, we skip Stage 2 and return the Stage 1 prediction. Otherwise, we proceed to Stage 2 with $[f_s,f_e]$ as the guidance interval.

\subsection{Stage 2: Attention-Guided Re-Inference}

When the confidence gate concludes that the Stage 1 prediction is unreliable, we ask the MLLM to regenerate its answer from the high-attention interval $[f_s,f_e]$. This directly addresses the perception-generation gap: the signal already captured by the TG-Heads can no longer be washed out by query-irrelevant frames during decoding.

\paragraph{Attention-guided crop.}
With a fixed total visual-token budget $B$ distributed across all $T$ frames, each frame receives on average $B/T$ tokens. By re-invoking the video processor on the cropped clip spanning the $T'{=}f_e{-}f_s{+}1$ frames in $[f_s, f_e]$, we obtain roughly $B/T'$ tokens per frame, giving the MLLM a finer visual resolution on the target segment. We rewrite all frame timestamps to start from zero to match the training distribution. The MLLM then produces a local prediction $[\hat{s}_{\text{loc}},\hat{e}_{\text{loc}}]$, which is mapped back to the original timeline by:
\begin{equation}
[\hat{s}^{(2)},\hat{e}^{(2)}] \;=\; \operatorname{clip}\!\left([\hat{s}_{\text{loc}},\; \hat{e}_{\text{loc}}] + (f_s-1)\Delta t,\; 0,\; D\right).
\end{equation}

\section{Experiments}

\begin{table*}[t]
    \caption{\textbf{Main Results.} ``+Ours'' denotes applying our inference-time framework to the backbone above. Gains are shown in green ($\uparrow$). Entries marked with $^{*}$ are taken from the TimeLens paper; the remaining results are evaluated on the same benchmark splits and metrics.}
    \centering
    \setlength{\tabcolsep}{1pt}
    \renewcommand{\arraystretch}{1.1}
    \resizebox{\textwidth}{!}{
    \begin{tabular}{l|cccc|cccc|cccc}
        \toprule
        \multicolumn{1}{c}{\multirow{2}{*}[-3pt]{\textbf{Model}}} 
        & \multicolumn{4}{c}{\textbf{QVHighlights-TimeLens}} 
        & \multicolumn{4}{c}{\textbf{Charades-TimeLens}} 
        & \multicolumn{4}{c}{\textbf{ActivityNet-TimeLens}} \\
        
        \cmidrule(lr){2-5} \cmidrule(lr){6-9} \cmidrule(lr){10-13}
        
        \multicolumn{1}{c}{}
        & \multicolumn{1}{c}{R1@0.3}
        & \multicolumn{1}{c}{R1@0.5}
        & \multicolumn{1}{c}{R1@0.7}
        & \multicolumn{1}{c}{mIoU}
        & \multicolumn{1}{c}{R1@0.3}
        & \multicolumn{1}{c}{R1@0.5}
        & \multicolumn{1}{c}{R1@0.7}
        & \multicolumn{1}{c}{mIoU}
        & \multicolumn{1}{c}{R1@0.3}
        & \multicolumn{1}{c}{R1@0.5}
        & \multicolumn{1}{c}{R1@0.7}
        & \multicolumn{1}{c}{mIoU} \\
        
        \midrule
        \multicolumn{1}{l}{\textit{Proprietary Models}} & & & & \multicolumn{1}{c}{}& & & & \multicolumn{1}{c}{} & & & & \\
        GPT-4o$^{*}$ & 69.0 & 54.8 & 38.5 & 52.1 & 60.6 & 44.5 & 23.5 & 41.8 & 55.2 & 41.4 & 25.8 & 40.4 \\
        GPT-5$^{*}$ & 72.4 & 60.4 & 46.4 & 56.8 & 59.3 & 42.0 & 22.0 & 40.5 & 57.4 & 44.9 & 30.4 & 42.9 \\
        Gemini-2.0-Flash$^{*}$ & 76.2 & 66.4 & 48.3 & 60.8 & 66.4 & 53.5 & 27.1 & 46.7 & 62.9 & 54.0 & 37.7 & 49.3 \\
        Gemini-2.5-Flash$^{*}$ & 78.2 & 69.4 & 55.0 & 64.3 & 68.7 & 56.1 & 30.6 & 48.6 & 66.8 & 57.5 & 41.3 & 52.5 \\
        Gemini-2.5-Pro$^{*}$ & 84.1 & 75.9 & 61.1 & 70.4 & 74.1 & 61.1 & 34.0 & 52.8 & 72.3 & 64.2 & 47.1 & 58.1 \\
        
        \midrule
        \multicolumn{1}{l}{\textit{Open-Source Models}} & & & & \multicolumn{1}{c}{}& & & & \multicolumn{1}{c}{} & & & & \\
        VideoChat-Flash-7B$^{*}$ & 45.2 & 30.6 & 16.7 & 32.7 & 60.2 & 37.9 & 17.8 & 39.7 & 35.5 & 21.8 & 10.5 & 24.8 \\
        VideoChat-R1-7B$^{*}$ & 29.3 & 19.1 & 9.4 & 21.5 & 51.9 & 30.8 & 11.7 & 33.7 & 35.0 & 23.9 & 11.3 & 25.0 \\
        Time-R1-7B$^{*}$ & 65.8 & 51.5 & 36.1 & 49.2 & 57.9 & 32.0 & 16.9 & 36.6 & 44.8 & 31.0 & 19.0 & 33.1 \\
        TimeSuite$^{*}$ & 27.1 & 16.9 & 9.9 & 21.7 & 56.3 & 35.5 & 18.0 & 38.1 & 27.1 & 17.5 & 8.6 & 19.8 \\
        Grounded-VideoLLM$^{*}$ & 43.7 & 33.8 & 22.5 & 33.4 & 43.3 & 28.7 & 13.5 & 30.0 & 39.2 & 29.6 & 19.5 & 30.0 \\
        Qwen2.5-VL-7B & 44.1 & 31.5 & 17.1 & 33.7 & 60.4 & 38.0 & 16.8 & 39.8 & 45.6 & 32.0 & 17.0 & 32.5 \\
        
        \midrule
        \rowcolor{black!5}
        
        MiMo-VL-7B & 65.1 & 55.4 & 40.9 & 51.8 & 63.8 & 46.5 & 24.4 & 44.4 & 58.3 & 45.8 & 29.2 & 42.6 \\
        
        \rowcolor{cyan!5}
        \textbf{MiMo-VL-7B (+Ours)}  
        & 70.1\up{5.0} 
        & 58.9\up{3.5} 
        & 43.5\up{2.6} 
        & 55.3\up{3.5} 
        & 65.0\up{1.2} 
        & 48.4\up{1.9} 
        & 25.9\up{1.5} 
        & 46.0\up{1.6} 
        & 59.6\up{1.3} 
        & 47.6\up{1.8} 
        & 30.4\up{1.2} 
        & 44.1\up{1.5} \\
        
        \rowcolor{black!5}
        Qwen3-VL-8B
        & 74.1 & 64.1 & 49.1 & 59.4 & 69.0 & 53.1 & 27.6 & 48.2 & 62.2 & 51.6 & 34.7 & 46.9 \\
        
        \rowcolor{cyan!5}
        \textbf{Qwen3-VL-8B (+Ours)} 
        & 77.3\up{3.2} 
        & 67.0\up{2.9} 
        & 51.2\up{2.1} 
        & 61.9\up{2.5} 
        & 71.2\up{2.2} 
        & 55.2\up{2.1} 
        & 28.3\up{0.7} 
        & 49.5\up{1.3} 
        & 64.2\up{2.0} 
        & 52.9\up{1.3} 
        & 35.4\up{0.7} 
        & 48.2\up{1.3} \\
        
        \rowcolor{black!5}
        Molmo2-8B
        & 79.4 & 69.0 & 53.9 & 63.3 & 59.7 & 46.1 & 24.1 & 41.4 & 64.9 & 53.8 & 36.0 & 48.7 \\

        \rowcolor{cyan!5}
        \textbf{Molmo2-8B (+Ours)}
        & \textbf{81.6}\up{2.2}
        & \textbf{72.9}\up{3.9}
        & \textbf{55.9}\up{2.0}
        & \textbf{65.7}\up{2.4}
        & \textbf{61.0}\up{1.3}
        & \textbf{48.1}\up{2.0}
        & \textbf{25.2}\up{1.1}
        & \textbf{42.7}\up{1.3}
        & \textbf{66.9}\up{2.0}
        & \textbf{57.1}\up{3.3}
        & \textbf{38.0}\up{2.0}
        & \textbf{50.6}\up{1.9} \\

        \bottomrule
    \end{tabular}}
    \label{tab:main_results}
\end{table*}

\subsection{Experimental Setup}
\label{sec:setup}

\textbf{Datasets and metrics.}
We evaluate on three benchmarks from TimeLens-Bench~\citep{zhang2025timelens}, which are manually re-annotated versions of three widely-used VTG datasets: \textbf{Charades-TimeLens}, built on Charades-STA~\citep{gao2017tall}; \textbf{ActivityNet-TimeLens}, built on ActivityNet Captions~\citep{krishna2017dense}; and \textbf{QVHighlights-TimeLens}, built on QVHighlights~\citep{lei2021detecting}. Together they cover a wide range of video domains, durations, and query semantics. Following standard practice~\citep{zhang2025timelens}, we report R1@$\theta$ (fraction of predictions with IoU $\geq\theta$) for $\theta\in\{0.3, 0.5, 0.7\}$ and the mean IoU (mIoU).

\textbf{Baselines.}
We apply our framework to three open-weight, general-purpose MLLMs that are not fine-tuned for VTG: \textbf{MiMo-VL-7B}~\citep{coreteam2025mimovltechnicalreport}, \textbf{Qwen3-VL-8B}~\citep{bai2025qwen3}, and \textbf{Molmo2-8B}~\citep{clark2026molmo2}. For broader comparison, Table~\ref{tab:main_results} also includes proprietary models GPT-4o~\citep{hurst2024gpt}, GPT-5~\citep{singh2025openai}, and the Gemini-2.x family~\citep{comanici2025gemini}, together with open-source VideoChat-Flash~\citep{li2024videochat}, VideoChat-R1~\citep{li2025videochat}, Time-R1~\citep{wang2025time}, TimeSuite~\citep{zeng2024timesuite}, Grounded-VideoLLM~\citep{wang2024grounded}, and Qwen2.5-VL~\citep{bai2025qwen2}.

\textbf{Implementation details.}
We strictly follow the inference protocol of TimeLens~\citep{zhang2025timelens}, including the per-video total visual-token budget, the frame sampling strategy, and the prompt template, so that our reproductions are directly comparable to the numbers reported in the TimeLens paper. Molmo2 is an exception: we evaluate it using its native video processor with 2 FPS sampling and up to 384 frames at $378\times378$ resolution, following its default video processing configuration. All backbones use the top-$5$ TG-Heads identified once per model by the GCS procedure in Section~\ref{sec:tghead}, after which they are fixed across the three benchmarks. All experiments use greedy decoding and run on 8$\times$H20 GPUs. Complete hyperparameter settings are in Appendix~\ref{sec:hyperparams}.

\subsection{Main Results}

\begin{table*}[t]
    \caption{\textbf{Qwen3-VL across model sizes} on QVHighlights-TimeLens. Gains are shown in green ($\uparrow$).}
    \centering
    \setlength{\tabcolsep}{1pt}
    \renewcommand{\arraystretch}{1.1}
    \resizebox{\textwidth}{!}{
    \begin{tabular}{l|cccc|cccc|cccc}
        \toprule
        \multicolumn{1}{c}{\multirow{2}{*}[-3pt]{\centering \textbf{Size}}} 
        & \multicolumn{4}{c}{\textbf{4B}} 
        & \multicolumn{4}{c}{\textbf{8B}} 
        & \multicolumn{4}{c}{\textbf{32B}} \\
        
        \cmidrule(lr){2-5} \cmidrule(lr){6-9} \cmidrule(lr){10-13}
        
        \multicolumn{1}{c}{}
        & \multicolumn{1}{c}{R1@0.3}
        & \multicolumn{1}{c}{R1@0.5}
        & \multicolumn{1}{c}{R1@0.7}
        & \multicolumn{1}{c}{mIoU}
        & \multicolumn{1}{c}{R1@0.3}
        & \multicolumn{1}{c}{R1@0.5}
        & \multicolumn{1}{c}{R1@0.7}
        & \multicolumn{1}{c}{mIoU}
        & \multicolumn{1}{c}{R1@0.3}
        & \multicolumn{1}{c}{R1@0.5}
        & \multicolumn{1}{c}{R1@0.7}
        & \multicolumn{1}{c}{mIoU} \\
        
        \midrule
        Qwen3-VL
        & 74.8 & 65.0 & 49.6 & 59.9 
        & 74.1 & 64.1 & 49.1 & 59.4 
        & 73.3 & 64.1 & 48.0 & 59.6 \\

        \textbf{+Ours}  
        & \textbf{76.1}\up{1.3} 
        & \textbf{66.0}\up{1.0} 
        & \textbf{50.4}\up{0.8} 
        & \textbf{61.0}\up{1.1} 
        & \textbf{77.3}\up{3.2} 
        & \textbf{67.0}\up{2.9} 
        & \textbf{51.2}\up{2.1} 
        & \textbf{61.9}\up{2.5} 
        & \textbf{75.4}\up{2.1} 
        & \textbf{65.9}\up{1.8} 
        & \textbf{50.5}\up{2.5} 
        & \textbf{62.0}\up{2.4} \\
        \bottomrule
    \end{tabular}}
    \label{tab:models_sizes}
\end{table*}

\begin{figure}[t]
  \centering
  \includegraphics[width=0.95\linewidth]{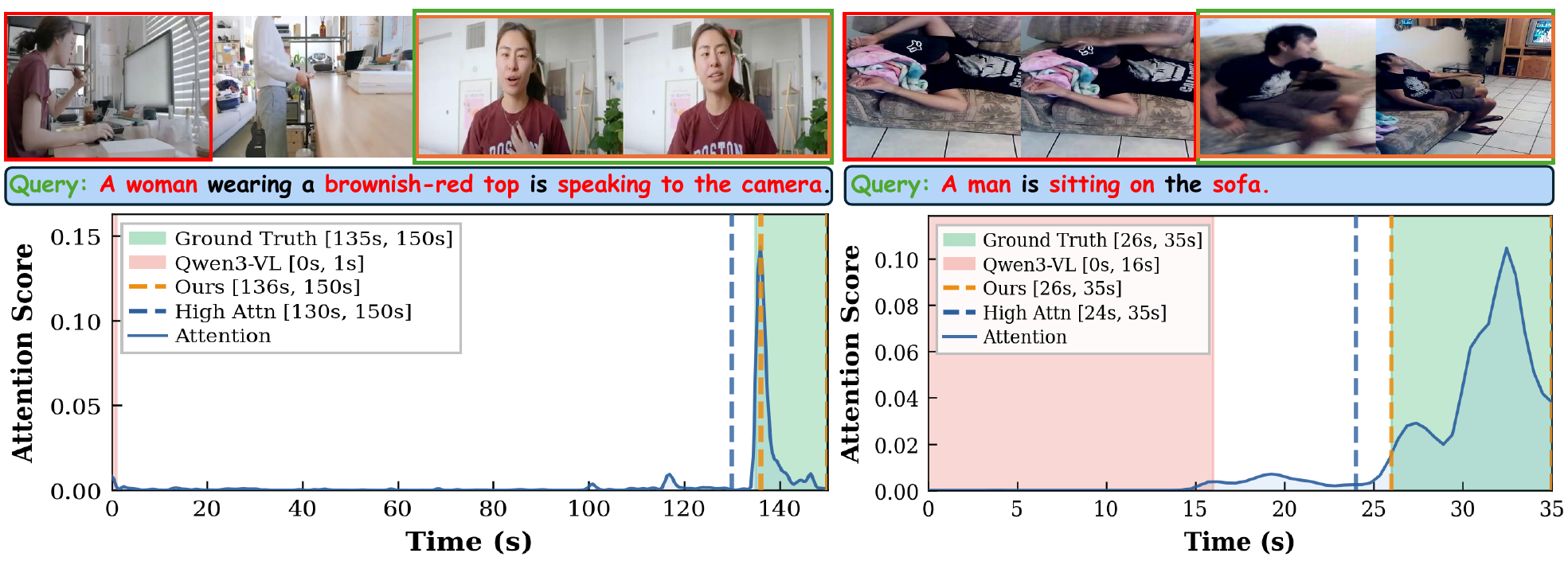}
  \caption{\textbf{Qualitative results.} Two examples where our method corrects an erroneous baseline prediction. Blue dashed boxes indicate the detected high-attention interval.}
  \label{fig:results}
  \vspace{-2mm}
\end{figure}

\textbf{Main results.}
Table~\ref{tab:main_results} summarizes the main results. Our framework delivers consistent improvements across all three general-purpose backbones and all three benchmarks. The gains are especially large on QVHighlights-TimeLens, where MiMo-VL-7B improves by $+3.5$ mIoU, Qwen3-VL-8B by $+2.5$ mIoU, and Molmo2-8B by $+2.4$ mIoU. Molmo2-8B also improves by $+1.3$ mIoU on Charades-TimeLens and $+1.9$ mIoU on ActivityNet-TimeLens, despite its already strong baseline. Overall, these results support our core claim that general-purpose MLLMs already possess a latent temporal grounding ability that can be surfaced without post-training. Results on the VTG-tuned TimeLens-8B model are reported in Appendix~\ref{sec:timelens_results}, and comparison with two representative training-free methods (Moment-GPT~\citep{xu2025zero}, NumPro~\citep{wu2025number}) can be found in Appendix~\ref{sec:training_free_comparison}.

\textbf{Models with Different Sizes.}
Table~\ref{tab:models_sizes} reports results for Qwen3-VL at three scales (4B, 8B, 32B). The performance gap across scales is surprisingly small for VTG (the 4B model at 59.9 mIoU is on par with the 32B model at 59.6 mIoU), and our framework delivers consistent gains at every scale, confirming that it serves as a plug-and-play enhancement regardless of backbone size.

\textbf{Qualitative evaluation.}
Figure~\ref{fig:results} visualizes two representative corrections. In the first case, a woman appears at the beginning of the video but is not speaking to the camera there. The baseline is misled by this visual similarity, while our TG-Head attention peaks at the correct later segment, and the detected high-attention interval (blue dashed boxes) excludes the distractor. In the second case, the man and sofa are present throughout the video, but the man is lying down earlier and only sits up later. The baseline confuses the two, while our high-attention interval filters out the earlier segment and guides Stage 2 to the correct prediction. Additional examples are in Appendix~\ref{sec:more_vis}.

\begin{table}[t]
\centering
\caption{\textbf{Ablation on contrastive attention debiasing} using Qwen3-VL-8B on three benchmarks.}
\label{tab:debiasing_ablation}
\setlength{\tabcolsep}{4pt}
\resizebox{\textwidth}{!}{
\begin{tabular}{l|cccc|cccc|cccc}
\toprule
\multicolumn{1}{c}{\multirow{2}{*}[-3pt]{\textbf{Method}}}
& \multicolumn{4}{c}{\textbf{QVHighlights-TimeLens}}
& \multicolumn{4}{c}{\textbf{Charades-TimeLens}}
& \multicolumn{4}{c}{\textbf{ActivityNet-TimeLens}} \\
\cmidrule(lr){2-5}\cmidrule(lr){6-9}\cmidrule(lr){10-13}
& R1@0.3 & R1@0.5 & R1@0.7 & mIoU
& R1@0.3 & R1@0.5 & R1@0.7 & mIoU
& R1@0.3 & R1@0.5 & R1@0.7 & mIoU \\
\midrule
w/o Debiasing & 75.4 & 65.0 & 49.3 & 60.4 & 69.6 & 53.8 & 28.0 & 48.6 & 62.8 & 51.8 & 34.9 & 47.5 \\
w/ Debiasing & \textbf{77.3} & \textbf{67.0} & \textbf{51.2} & \textbf{61.9}
                    & \textbf{71.2} & \textbf{55.2} & \textbf{28.3} & \textbf{49.5}
                    & \textbf{64.2} & \textbf{52.9} & \textbf{35.4} & \textbf{48.2} \\
\bottomrule
\end{tabular}}
\end{table}

\begin{figure}[t]
  \centering
  \includegraphics[width=1.0\linewidth]{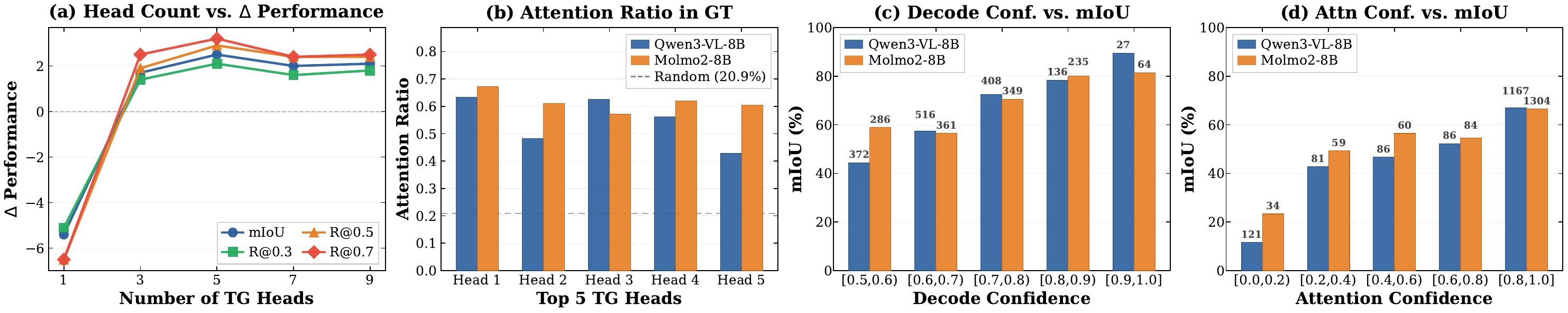}
  \caption{\textbf{Analysis on Qwen3-VL-8B and Molmo2-8B} (QVHighlights-TimeLens). (a)~Performance gain vs.\ number of TG-Heads $K$. (b)~Attention ratio of the ground-truth interval in each of the top-5 TG-Heads, compared to a random baseline (dashed line). (c, d)~Stage~1 mIoU bucketed by decode confidence (c) and attention confidence (d); numbers above bars are sample counts.}
  \label{fig:analysis}
\end{figure}

\subsection{More Analysis}
\label{sec:analysis}

\textbf{Sensitivity to Attention Heads.}
Figure~\ref{fig:analysis}(a) shows performance as a function of the number of TG-Heads $K$. Using a single head ($K{=}1$) actually degrades performance, as individual heads lack sufficient coverage to reliably localize diverse events. Performance improves steadily as more heads are added, peaking at $K{=}5$, where multiple complementary heads jointly identify the optimal interval. Beyond $K{=}5$, weaker heads introduce noise that broadens the detected interval to include irrelevant regions, causing a slight decline. We therefore use $K{=}5$ for all experiments.

\textbf{Do MLLMs Really Know When to Look?} Figure~\ref{fig:analysis}(b) provides direct evidence. We compute the attention ratio, defined as the fraction of total attention falling within the ground-truth interval, for the top-5 TG-Heads. This ratio consistently exceeds random for both Qwen3-VL-8B and Molmo2-8B. Moreover, the prefill-derived interval alone reaches 51.7 mIoU on QVHighlights (Appendix~\ref{sec:attention_interval_only}), showing that useful temporal localization is already present before decoding.


\textbf{The Effect of Confidence Gate.}
Figures~\ref{fig:analysis}(c,d) show that both decode confidence and attention confidence are positively correlated with mIoU. In the majority of cases, the model is confident and achieves high accuracy, indicating that its Stage 1 prediction is already reliable. However, a non-negligible fraction of samples fall into low-confidence bins, and these samples consistently correspond to poor localization. These are precisely the cases where the model's decoding drifts away from its own attention evidence. Our confidence gate identifies them and routes them to Stage 2 for correction, while leaving the already-accurate high-confidence predictions untouched.

\textbf{Ablation on Contrastive Attention Debiasing.}
Table~\ref{tab:debiasing_ablation} compares our method with and without contrastive attention debiasing on Qwen3-VL-8B across all three benchmarks. Without debiasing, the detected intervals are frequently biased toward the beginning of the video due to the attention sink effect, leading to degradation across all metrics. Contrastive debiasing effectively alleviates this bias, resulting in more accurate interval detection and consistently better performance.

\begin{wraptable}{r}{0.48\textwidth}
    \vspace{-0.45cm} 
    \captionsetup{font=small} 
    \caption{\textbf{Ablation on attention aggregation strategies} (Qwen3-VL-8B, QVHighlights-TimeLens). }
    \vspace{-0.1cm} 
    \centering

    \resizebox{0.48\columnwidth}{!}{
    \begin{tabular}{lcccc}
    \toprule
    \textbf{Method}  & R1@0.3  & R1@0.5  & R1@0.7  & mIoU \\
    \midrule
    Baseline   & 74.1 & 64.1 & 49.1 & 59.4 \\
    \midrule
    Average   & 75.5 & 64.2 & 48.5 & 59.8 \\
    Next Token & 77.0 & 66.6 & 50.6 & 61.5 \\
    Last Token & 75.2 & 64.9 & 51.0 & 61.2 \\
    Entropy (Ours)    & \textbf{77.3} & \textbf{67.0} & \textbf{51.2} & \textbf{61.9} \\
    \bottomrule
    \end{tabular}
    }
    \label{tab:ablation}
    \vspace{-0.6cm} 
\end{wraptable}

\textbf{Ablations on Attention Aggregation.}
Table~\ref{tab:ablation} compares four aggregation strategies on Qwen3-VL-8B. \emph{Average} treats all query tokens equally, yielding marginal gains (+0.4 mIoU) because uninformative tokens dilute the temporal signal. \emph{Next Token} uses the attention of the token immediately following the query, which benefits from causal self-attention over the full query context and achieves strong results (+2.1 mIoU). \emph{Last Token} uses the final prompt token to represent the holistic query semantics. Our \emph{entropy-based} aggregation achieves the best overall performance (+2.5 mIoU) by automatically identifying and upweighting the most temporally discriminative query tokens without requiring positional heuristics.

\section{Conclusion}

We have shown that MLLMs often know \emph{when} before they speak. By probing query-to-video attention, we identify Temporal Grounding Heads whose prefill attention concentrates on the target interval, even when the final generated timestamp is incorrect. This reveals a perception-generation gap: temporal grounding evidence is present inside the model, but can be diluted or redirected during decoding. Our inference-time read-then-regenerate framework recovers this hidden signal by converting TG-Head attention into a temporal relevance cue and using it to restrict distracting visual context before regeneration. Improvements across various MLLMs demonstrate that bridging the gap between what MLLMs internally attend to and what they ultimately generate can enhance VTG without modifying model parameters, and may benefit broader reliable multimodal reasoning.


\section*{AI Use Statement}
Generative AI tools were used only for language polishing and editing of the manuscript, not as a component of the research methodology. The authors reviewed all AI-assisted text and take responsibility for the final content of this work, including its claims and presentation.

\section*{Reproducibility Statement}
Section~\ref{sec:setup} describes the datasets, baselines, metrics, and evaluation setup. Appendix~\ref{sec:impl_details_appendix} provides the inference protocol and implementation details, Appendix~\ref{sec:hyperparams} reports the hyperparameter settings, and Appendix~\ref{sec:efficiency} analyzes computational efficiency. These sections together document the information needed to reproduce the reported results.

\bibliography{iclr2027_conference}
\bibliographystyle{iclr2027_conference}


\newpage
\appendix

\startcontents[appendices]

\section*{Appendix Contents}
\printcontents[appendices]{}{1}{
    \setcounter{tocdepth}{2}
}

\section{Additional Evidence for the Perception--Generation Gap}
\label{app:gap}
\subsection{Additional Motivation Examples}
\label{sec:motivation_appendix}

\begin{figure}[ht]
  \centering
  \includegraphics[width=0.9\linewidth]{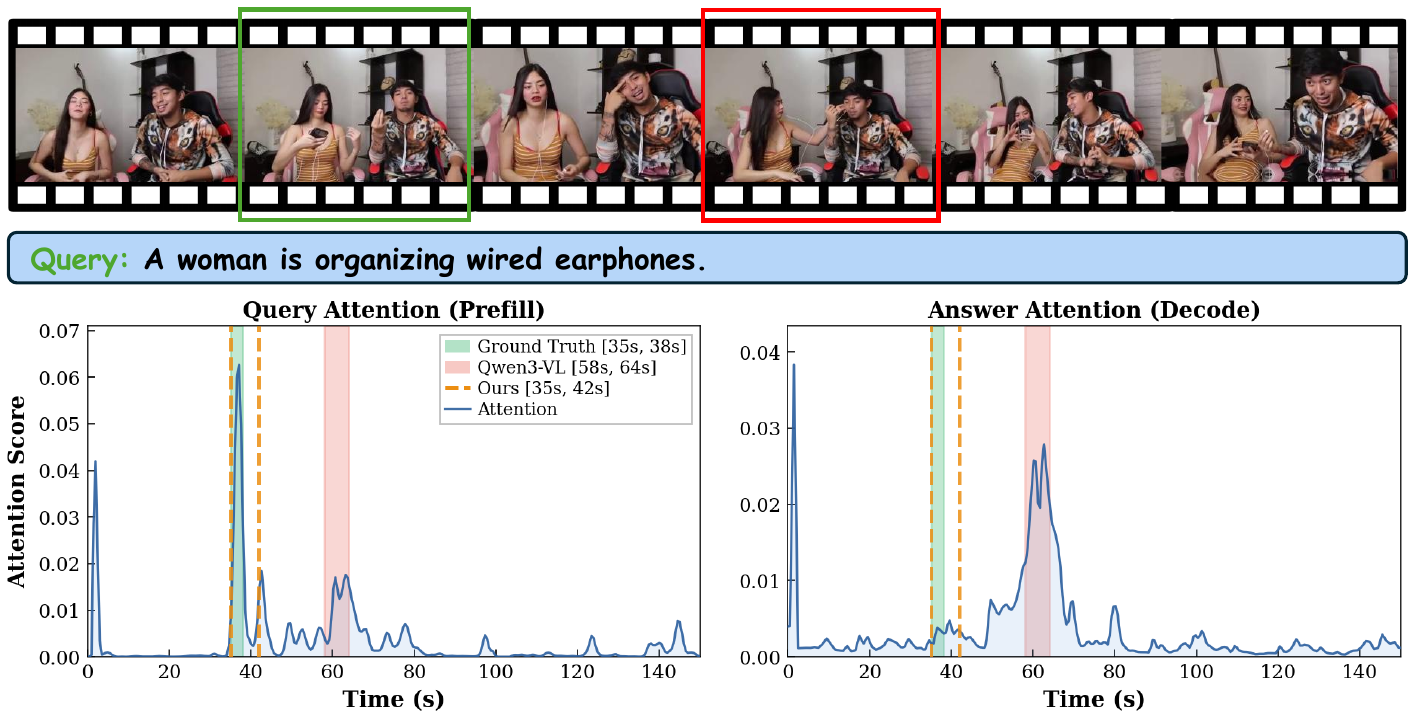}
  \caption{Additional motivation example. Query attention (prefill) correctly focuses on the ground-truth interval, while answer attention (decoding) is distracted by a visually similar but query-irrelevant later segment.}
  \label{fig:motivation_1_appendix}
\end{figure}

\begin{figure}[ht]
  \centering
  \includegraphics[width=0.9\linewidth]{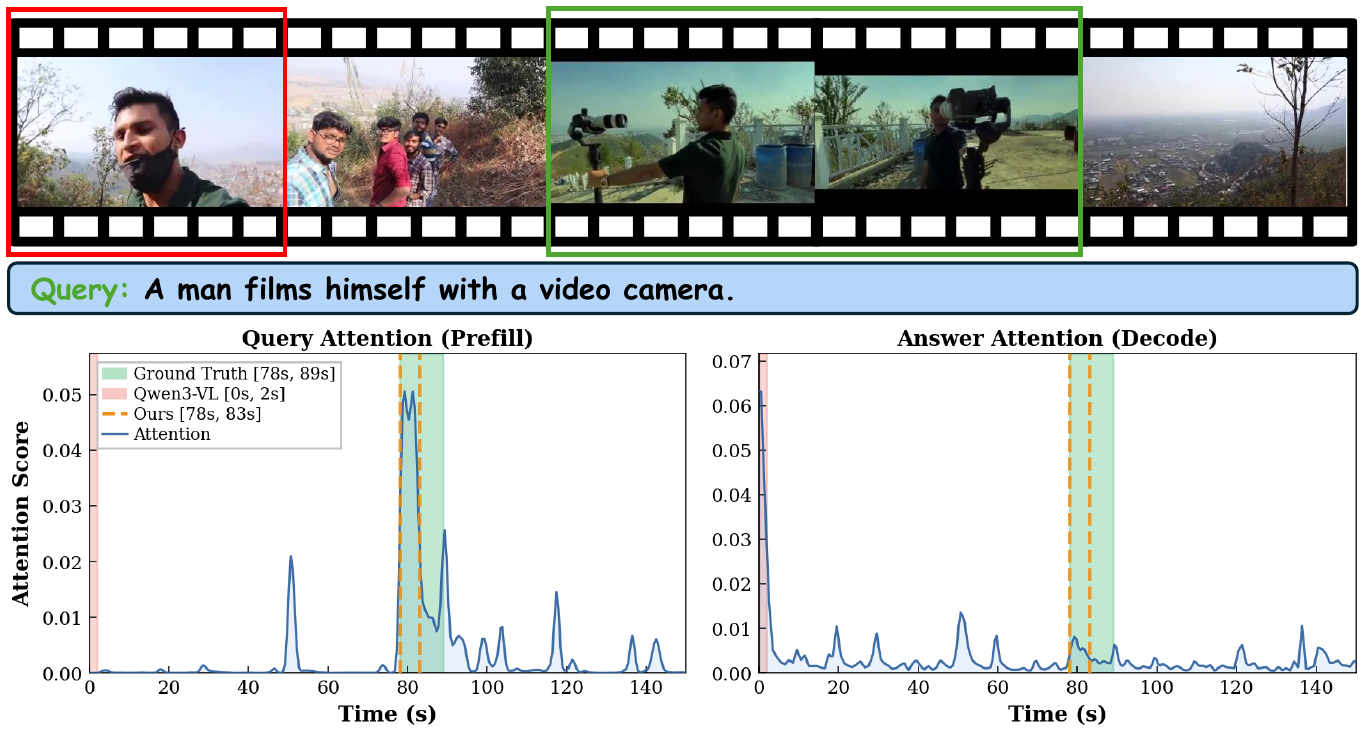}
  \caption{Additional motivation example. Query attention (prefill) sharply peaks at the ground-truth interval, but answer attention (decoding) is overwhelmed by the attention sink at the beginning of the video, causing the baseline to mispredict.}
  \label{fig:motivation_2_appendix}
\end{figure}

To supplement Figure~\ref{fig:motivation} in the main paper, we provide two additional motivating examples in Figures~\ref{fig:motivation_1_appendix} and~\ref{fig:motivation_2_appendix}. Both examples reinforce the central observation that MLLMs reliably localize the queried event during prefill but drift away from this localization during autoregressive decoding.

In Figure~\ref{fig:motivation_1_appendix}, the query asks about a woman organizing wired earphones. The query attention (prefill) sharply peaks around the ground-truth interval [35s, 38s], and our method successfully predicts [35s, 42s]. In contrast, the answer attention (decoding) drifts to a later segment around [58s, 64s] where the scene is visually similar but the target action is not actually occurring, and the baseline Qwen3-VL predicts exactly this wrong interval.

Figure~\ref{fig:motivation_2_appendix} shows a case where the answer attention is overwhelmed by the very first frame rather than by a distractor segment. The query is about a man filming himself with a video camera. While the query attention still clearly localizes the ground-truth interval [78s, 89s], the answer attention is dominated by the attention sink at the beginning of the video, and the model ends up predicting [0s, 2s]. Our method produces [78s, 83s], closely matching the ground truth.

\subsection{How Informative Is the Prefill Attention Interval?}
\label{sec:attention_interval_only}

Our central claim is not that the high-attention interval can replace the MLLM's timestamp decoder, but that the prefill attention already contains a useful coarse localization signal before autoregressive generation begins. We therefore directly evaluate the interval $[f_s,f_e]$ extracted from TG-Head prefill attention as a standalone VTG prediction, without Stage~2 regeneration.

\begin{table}[ht]
\centering
\caption{\textbf{Standalone localization accuracy of the TG-Head prefill interval} on Qwen3-VL-8B, QVHighlights-TimeLens. The attention interval is designed as a recall-oriented guidance region for Stage~2 rather than a final boundary prediction.}
\label{tab:attention_interval_only}
\begin{tabular}{lcccc}
\toprule
\textbf{Prediction} & R1@0.3 & R1@0.5 & R1@0.7 & mIoU \\
\midrule
MLLM decoding (baseline) & 74.1 & 64.1 & 49.1 & 59.4 \\
TG-Head prefill interval only & 72.3 & 58.2 & 37.6 & 51.7 \\
Read-then-regenerate (Ours) & \textbf{77.3} & \textbf{67.0} & \textbf{51.2} & \textbf{61.9} \\
\bottomrule
\end{tabular}
\end{table}

The prefill interval alone already provides substantial temporal localization: it reaches 72.3 R1@0.3 without using the MLLM's generated timestamp prediction. At the same time, it is less precise than direct decoding at stricter IoU thresholds. This behavior is expected because the interval detector is deliberately recall-oriented: thresholding, dilation, and boundary expansion are designed to retain sufficient surrounding context and reduce the risk of excluding the target event before Stage~2. Thus, the TG-Head interval is not intended to replace the MLLM's decoding ability. Instead, it provides a coarse but informative temporal prior that can be combined with the model's stronger boundary-generation capability during re-inference.

\subsection{Dataset-Level Prefill--Decoding Gap}
\label{sec:prefill_decoding_gap}

The qualitative examples in
Figures~\ref{fig:motivation},
\ref{fig:motivation_1_appendix}, and
\ref{fig:motivation_2_appendix}
suggest that useful temporal evidence can already emerge during prefill even when the final decoded timestamp is incorrect. We next quantify how frequently this phenomenon occurs at the dataset level.

A direct IoU criterion is not ideal for evaluating the prefill-derived interval
$I_{\mathrm{pre}}=[f_s,f_e]$, because this interval is intentionally constructed as a recall-oriented guidance region rather than a precise boundary prediction. In particular, the thresholding, dilation, and boundary-expansion steps are designed to preserve sufficient surrounding context and reduce the risk of excluding the target event before Stage~2. Consequently, the prefill interval may have modest IoU with the ground truth even when it retains nearly the entire queried event.

We therefore measure the \emph{ground-truth coverage} of the prefill interval:
\begin{equation}
\mathrm{Cov}_{\mathrm{GT}}(I_{\mathrm{pre}},G)
=
\frac{|I_{\mathrm{pre}}\cap G|}{|G|},
\end{equation}
where $G$ denotes the ground-truth interval. We consider the target event to be successfully retained by the prefill signal when
$\mathrm{Cov}_{\mathrm{GT}}(I_{\mathrm{pre}},G) > 0.95$.
A Stage~1 decoding failure, denoted by $\mathrm{Dec}\times$, is defined as before: the decoded prediction has temporal IoU below 0.3 with the ground truth.

We report two complementary statistics. First,
$P(\mathrm{Pre}\checkmark \mid \mathrm{Dec}\times)$
measures, among Stage~1 decoding failures, the fraction for which the prefill-derived interval still retains more than 95\% of the ground-truth event. Second,
$P(\mathrm{Pre}\checkmark,\mathrm{Dec}\times)$
measures the prevalence of such prefill-retained but decode-wrong cases over the full evaluation set.

High ground-truth coverage could, however, be achieved trivially by retaining most of the video. To quantify how selective the prefill guidance is, we additionally report the \emph{guidance ratio}, defined as the duration of the prefill-derived interval divided by the full video duration.

\begin{table}[ht]
\centering
\caption{
Dataset-level prefill--decoding gap.
$\mathrm{Pre}\checkmark$ denotes that the prefill-derived guidance interval covers more than 95\% of the ground-truth event, while $\mathrm{Dec}\times$ denotes that the Stage~1 decoded prediction has IoU $<0.3$ with the ground truth.
The guidance ratio is the average duration of the prefill interval relative to the full video.
}
\label{tab:prefill_decode_gap}
\begin{tabular}{llccc}
\toprule
Model & Benchmark &
$P(\mathrm{Pre}\checkmark \mid \mathrm{Dec}\times)$ &
$P(\mathrm{Pre}\checkmark,\mathrm{Dec}\times)$ &
Guidance Ratio \\
\midrule
Qwen3-VL-8B & QVHighlights & 98\% & 25\% & 57\% \\
Qwen3-VL-8B & Charades     & 96\% & 30\% & 48\% \\
Qwen3-VL-8B & ActivityNet  & 95\% & 36\% & 58\% \\
Molmo2-8B    & QVHighlights & 99\% & 20\% & 74\% \\
MiMo-VL-7B   & QVHighlights & 98\% & 34\% & 54\% \\
\bottomrule
\end{tabular}
\end{table}

As shown in Table~\ref{tab:prefill_decode_gap}, the prefill-derived guidance interval retains nearly the entire target event in the overwhelming majority of decoding failures. Across all evaluated settings, 95--99\% of Stage~1 failures still have a prefill interval covering more than 95\% of the ground-truth event. Such cases are also common at the dataset level: prefill-retained but decode-wrong examples account for approximately 20--36\% of all evaluation samples.

Importantly, this high coverage is not obtained by simply retaining the entire video. The prefill guidance interval occupies on average 48--74\% of the original video duration, removing roughly 26--52\% of the visual context while still preserving almost the complete target event in nearly all decoding failures. This indicates that the prefill attention provides a substantially more focused temporal region than the full video, even though its boundaries are deliberately conservative.

These results clarify the nature of the perception--generation gap. The prefill-derived interval is not intended to recover exact temporal boundaries. Rather, it serves as a coarse, high-recall temporal prior that identifies a region containing the queried event. Even when autoregressive decoding produces an incorrect temporal span, the relevant event has typically already been retained within the prefill-guided region. Stage~2 can then operate on this reduced visual context to recover more precise start and end boundaries.

Taken together, these results suggest that many VTG errors do not arise from a complete failure to identify where the queried event occurs. Instead, useful temporal evidence is often already present during prefill but is not faithfully reflected in the final decoded timestamps, motivating our read-then-regenerate framework.

\subsection{Decode-Only Attention Steering}
\label{sec:decode_steering}

The perception-generation gap reported in the main paper is observational: TG-Head attention localizes the target during prefill but drifts away during decoding. A natural concern is whether this drift merely correlates with the final error or actually causes it. To test causality, we perform a decode-only intervention that corrects the decoding-stage attention without otherwise changing the model's input, visual encoding, or answer format.

\paragraph{Setup.}
We use the same Stage~1 pipeline to obtain the detected interval $[f_s,f_e]$, but instead of cropping and re-running the model, we keep the original video and prompt unchanged and bias only the TG-Head attention during autoregressive decoding: at every decoding step we add a constant to the attention logits that up-weights visual tokens inside $[f_s,f_e]$ and down-weights those outside it. No video is re-read or re-encoded, no second prefill is performed, and the predicted span is produced by the same greedy decoding process as the baseline. The only difference from the baseline is therefore where the answer tokens look while they are being generated.

\paragraph{Result.}
Table~\ref{tab:decode_only} reports mIoU on QVHighlights-TimeLens with Qwen3-VL-8B. Merely redirecting decoding-stage attention back to the interval that the model itself identified during prefill improves mIoU from $59.4$ to $60.9$, recovering $60\%$ of the gain of the full read-then-regenerate framework ($61.9$). This shows that decoding-stage drift is a causal contributor rather than a passive correlate: a substantial part of the mistake is made while the model is \emph{speaking}, not while it is \emph{reading}.

\begin{table}[ht]
\centering
\caption{\textbf{Decode-only attention steering} (Qwen3-VL-8B, QVHighlights-TimeLens). Correcting only where the answer tokens attend during decoding recovers a large part of the improvement, indicating that decoding drift is causal rather than merely correlational.}
\label{tab:decode_only}
\setlength{\tabcolsep}{4pt}
\begin{tabular}{lcccc}
\toprule
\textbf{Variant} & \textbf{R1@0.3} & \textbf{R1@0.5} & \textbf{R1@0.7} & \textbf{mIoU} \\
\midrule
Base Model & 74.1 & 64.1 & 49.1 & 59.4 \\
Decode-only attention steering& 75.3 & 65.9 & 50.7 & 60.9 \\
Ours & \textbf{77.3} & \textbf{67.0} & \textbf{51.2} & \textbf{61.9} \\
\bottomrule
\end{tabular}
\end{table}

\section{Analysis of Temporal Grounding Heads}

\subsection{Effectiveness of TG-Heads Selection}
\label{sec:tghead_effectiveness}

To validate that the TG-Heads identified by our knockout procedure are indeed the right heads to use, we compare several alternative head-selection strategies on Qwen3-VL-8B (QVHighlights-TimeLens) in Table~\ref{tab:tghead_ablation}. All variants use the same downstream pipeline (entropy aggregation, contrastive debiasing, interval detection, and Crop re-inference); only the set of heads from which attention is extracted differs.

\begin{table}[ht]
\centering
\caption{\textbf{Ablation on head selection} (Qwen3-VL-8B, QVHighlights-TimeLens). ``All heads'' extracts attention from every head in the model. ``Random 5'' averages five randomly sampled heads (mean over 3 draws). ``Top 5--10'' uses the heads ranked 6th--10th by GCS. ``Top 5 (Ours)'' uses the five highest-GCS heads.}
\label{tab:tghead_ablation}
\begin{tabular}{lcccc}
\toprule
\textbf{Head Selection} & R1@0.3 & R1@0.5 & R1@0.7 & mIoU \\
\midrule
Baseline & 74.1 & 64.1 & 49.1 & 59.4 \\
\midrule
All heads              & 72.5 & 62.2 & 47.6 & 57.5 \\
Random 5 heads         & 73.1 & 63.5 & 47.9 & 58.3 \\
Top 5--10 heads        & 75.7 & 65.8 & 50.1 & 60.6 \\
Top 5 heads (Ours)     & \textbf{77.3} & \textbf{67.0} & \textbf{51.2} & \textbf{61.9} \\
\bottomrule
\end{tabular}
\end{table}

Two observations stand out. First, using all heads or a random subset of heads actually \emph{degrades} performance relative to the baseline (e.g., All heads: 57.5 vs.\ Baseline: 59.4 mIoU). This confirms that the vast majority of heads do not carry useful temporal grounding information. Including them injects noise that corrupts the attention signal and misleads the interval detection. Second, the top-5 heads selected by GCS substantially outperform the next-best tier (Top 5--10), which in turn outperforms the random and all-head baselines. This validates the head knockout procedure as an effective way to isolate the most informative heads and demonstrates that TG-Head selection is critical to the success of our framework.

\subsection{Calibration Size and Stability}
\label{sec:calibration_stability}

\paragraph{How many calibration samples are needed?}
The TG-Heads are selected once per backbone using labeled calibration examples, and are then fixed across all evaluation benchmarks. To examine whether the selected heads depend strongly on the calibration-set size, we repeat the complete head-selection and threshold-calibration procedure using only 50 and 200 examples for Qwen3-VL-8B on QVHighlights-TimeLens.

\begin{table}[ht]
\centering
\caption{\textbf{Sensitivity to calibration-set size} for Qwen3-VL-8B on QVHighlights-TimeLens. TG-Heads and confidence thresholds are re-selected independently for each calibration size.}
\label{tab:calibration_size}
\begin{tabular}{lcccc}
\toprule
\textbf{\# Calibration Samples} & R1@0.3 & R1@0.5 & R1@0.7 & mIoU \\
\midrule
50  & 76.7 & 66.9 & 50.8 & 61.5 \\
200 & 77.3 & 67.0 & 51.2 & 61.9 \\
500 (default) & 77.3 & 67.0 & 51.2 & 61.9 \\
\bottomrule
\end{tabular}
\end{table}

The selection is highly stable once roughly 200 calibration examples are available. With 200 samples, the selected top-5 TG-Heads are identical to those obtained with the default 500-sample calibration set. With only 50 samples, the selected heads are
$(15,11)$, $(15,8)$, $(23,4)$, $(18,15)$, and $(21,10)$; all five remain within the top-8 heads ranked using 500 samples, and the downstream mIoU decreases by only 0.4 points. These results indicate that the calibration primarily needs enough examples to reliably rank the same small set of high-contribution heads, rather than hundreds of examples.

\paragraph{Role of ground-truth timestamps.}
For each backbone, we use the same 500 labeled calibration examples for both
TG-Head ranking and confidence-threshold calibration. These examples are
randomly sampled from the TimeLens-100K training corpus and are disjoint from
all evaluation benchmarks. Ground-truth timestamps are used only during this
one-time calibration stage; no model parameters are updated during calibration or inference. The resulting TG-Heads and confidence thresholds are fixed and reused across all three benchmarks. We use 500 examples by default, while Table~\ref{tab:calibration_size} shows that 200 examples recover the same top-5 heads and downstream performance.

\subsection{Long-Tailed Distribution of TG-Heads}
\label{sec:mIOU_drop_appendix}

\begin{figure}[ht]
  \centering
  \includegraphics[width=0.95\linewidth]{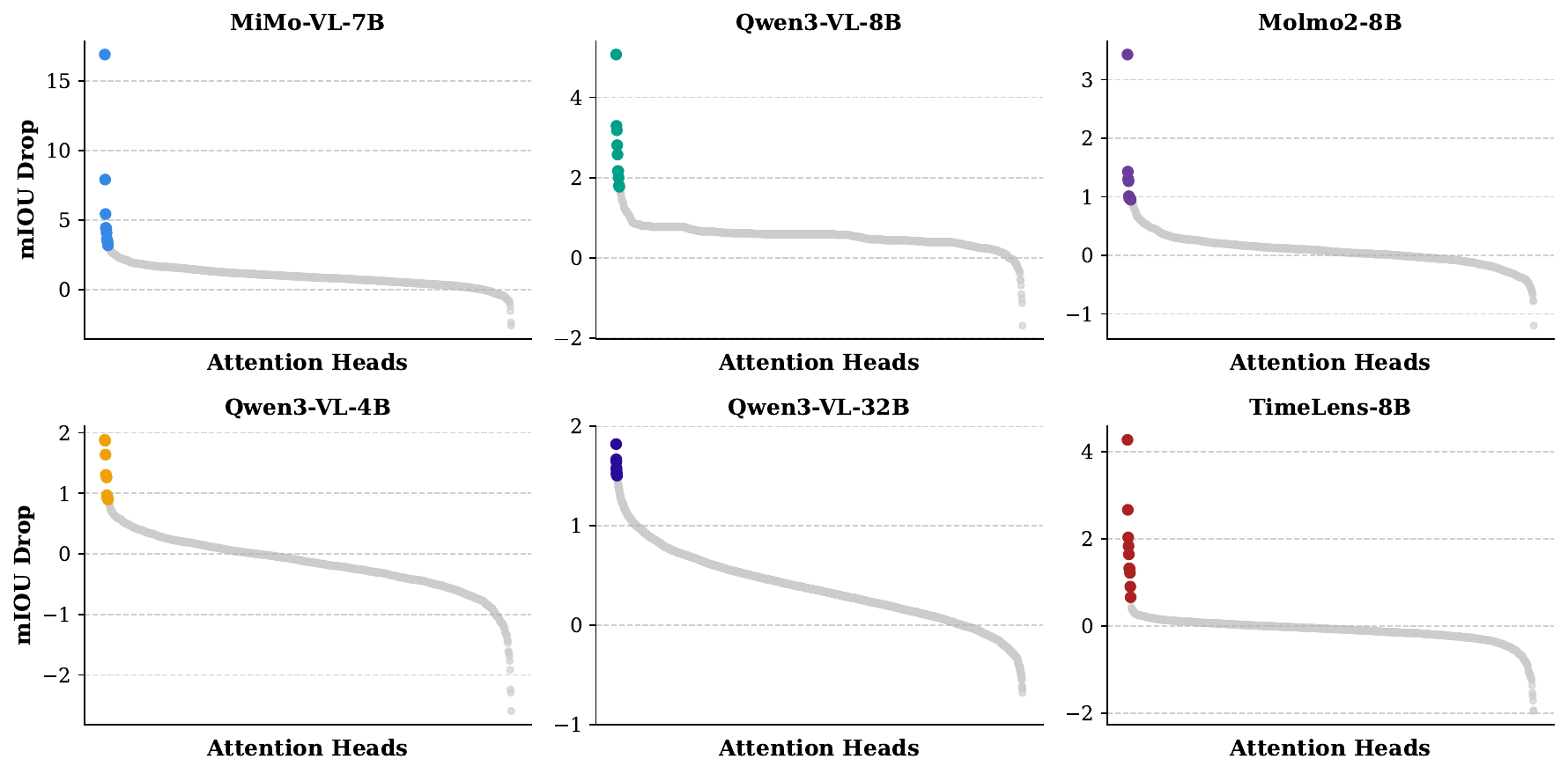}
  \caption{\textbf{Sorted mIoU drop curves under single-head pruning} for six MLLMs. The drop is heavy-tailed: a small subset of heads has a disproportionate effect on grounding performance, while the remaining heads are largely redundant or noisy.}
  \label{fig:miou_drop}
\end{figure}

Figure~\ref{fig:miou_drop} shows the sorted mIoU drops obtained when knocking out each individual head for six MLLMs. For every model, the curve exhibits a steep initial decline followed by a long flat tail that may yield slight gains at the extreme end. This confirms that the heavy-tailed concentration of temporal grounding capability is not specific to a single architecture or model scale: MiMo-VL-7B, Molmo2-8B, Qwen3-VL at 4B, 8B, and 32B scales, and TimeLens-8B all rely disproportionately on a small subset of heads, while the vast majority are redundant or inert. This long-tailed pattern justifies selecting only the top-$K$ heads, with $K$ as small as 5.

\subsection{TG-Heads Across Different Models}
\label{sec:tg_heads_models}

\begin{table}[ht]
\centering
\caption{Five TG-Head indices used for each model.}
\label{tab:tg_heads_indices}
\begin{tabular}{lc}
\toprule
\textbf{Model} & \textbf{TG-Head Indices (layer, head)} \\
\midrule
MiMo-VL-7B & (19, 3), (20, 19), (14, 17), (0, 1), (28, 24) \\
Molmo2-8B & (15, 11), (21, 10), (21, 11), (16, 23), (21, 31) \\
TimeLens-8B & (15, 8), (18, 22), (15, 0), (15, 2), (17, 24) \\
Qwen3-VL-4B & (21, 11), (27, 18), (18, 14), (20, 15), (28, 16) \\
Qwen3-VL-8B & (15, 11), (15, 8), (18, 15), (21, 31), (29, 12) \\
Qwen3-VL-32B & (19, 54), (43, 52), (23, 40), (33, 60), (2, 40) \\
\bottomrule
\end{tabular}
\end{table}

\begin{figure}[ht]
  \centering
  \includegraphics[width=0.95\linewidth]{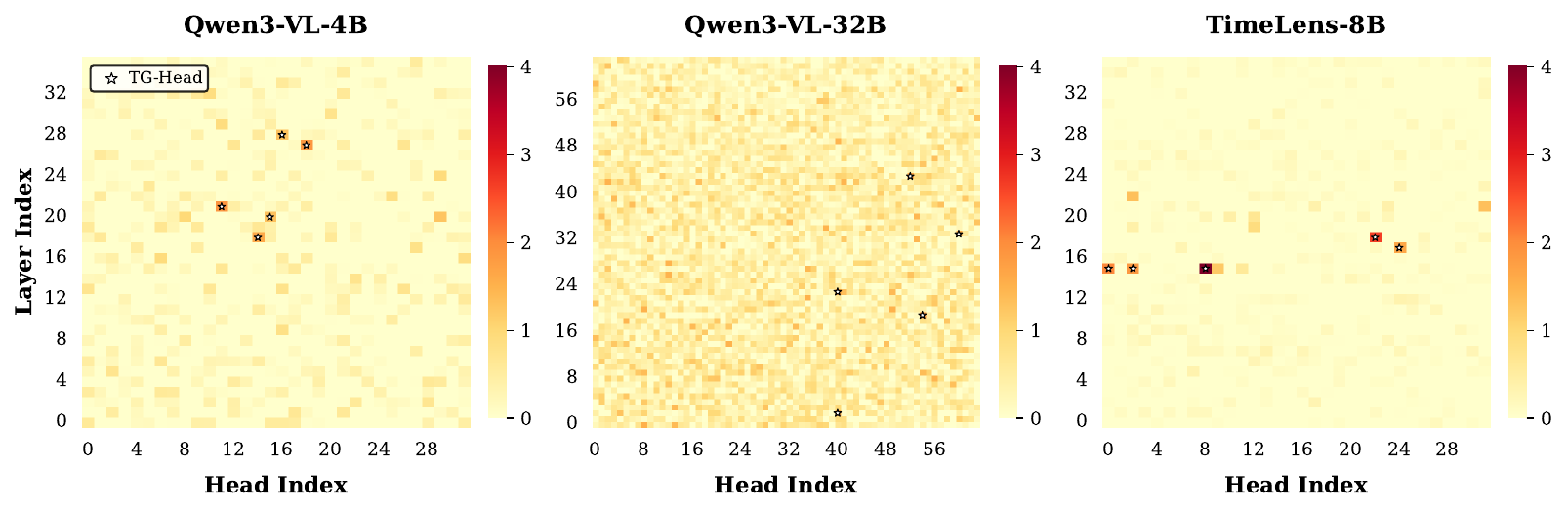}
  \caption{\textbf{Head-level grounding scores for Qwen3-VL-4B, Qwen3-VL-32B, and TimeLens-8B.} Stars mark the five selected TG-Heads. The sparse concentration pattern generalizes across model scales and to the VTG-tuned TimeLens backbone.}
  \label{fig:TG_heads_4-32b}
\end{figure}

Table~\ref{tab:tg_heads_indices} lists the top-5 GCS heads used for each evaluated model. Although most selected heads occur in middle-to-deep layers, a few shallow heads also contribute, indicating that temporal grounding can draw on both positional and high-level cross-modal signals. The selected heads remain fixed across datasets after one-time model-level calibration.

Figure~\ref{fig:TG_heads_4-32b} reports the head-level score heatmaps for Qwen3-VL-4B, Qwen3-VL-32B, and TimeLens-8B, which are omitted from the main paper for space. All three exhibit a sparse concentration pattern, and the selected heads are marked with stars.

\begin{figure}[t]
  \centering
  \includegraphics[width=0.9\linewidth]{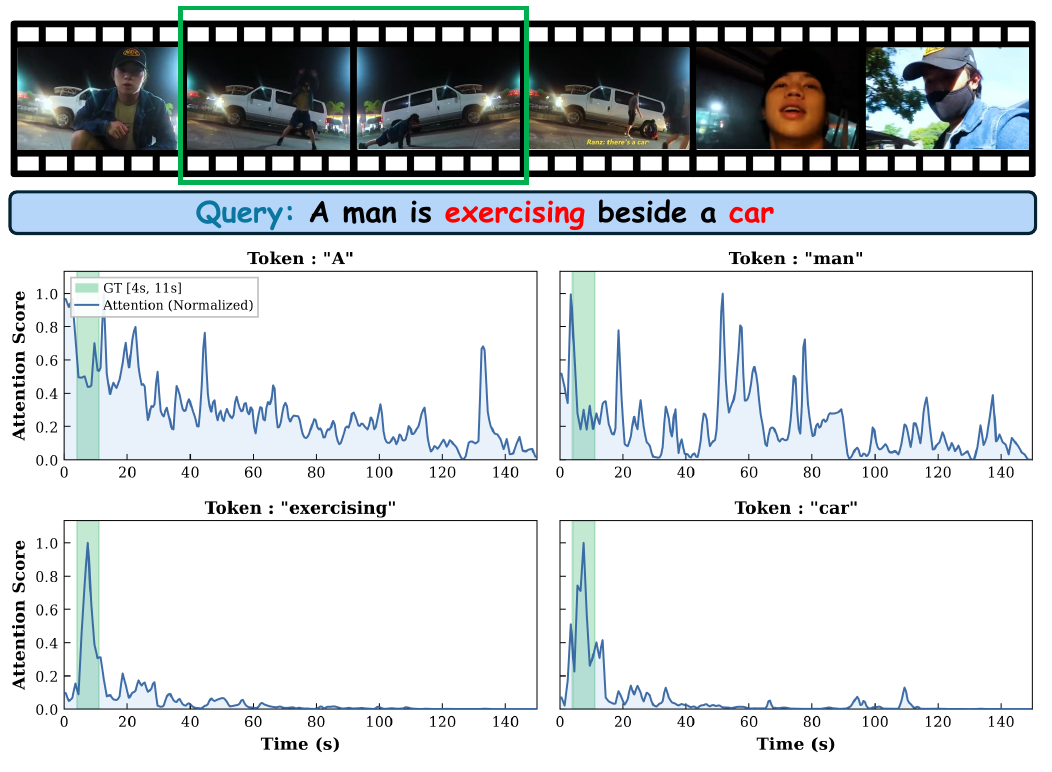}
  \caption{\textbf{Per-token attention distributions.} Function and generic tokens (``A'', ``man'') yield high-entropy, uninformative curves, while content tokens (``exercising'', ``car'') yield low-entropy curves that sharply peak inside the ground-truth interval. Entropy-based weighting prioritizes these informative tokens.}
  \label{fig:entropy_example}
\end{figure}

\section{Method Component Analysis}

\subsection{Entropy-Based Attention Aggregation}
\label{sec:entropy_appendix}

The query sentence contains tokens with very different temporal discriminativeness. Figure~\ref{fig:entropy_example} visualizes the per-token attention curves of a single TG-Head for the query ``\emph{A man is exercising beside a car}''. Function tokens such as ``\emph{A}'' distribute attention almost uniformly across the entire video, producing a high-entropy curve that is dominated by the attention sink at the very beginning and by generic background noise. Tokens that repeatedly appear across frames such as ``\emph{man}'' are also noisy: the model attends to multiple people in the video, not only the target man. In contrast, content tokens that are specific to the queried action or object, such as ``\emph{exercising}'' and ``\emph{car}'', produce sharp peaks inside the ground-truth interval (low entropy), accurately highlighting the event of interest.

A naive approach that simply averages all per-token attention curves would be dominated by the noisy, high-entropy tokens that contribute most of the energy but none of the temporal discriminativeness. Our entropy-based weighting inversely prioritizes low-entropy (i.e., concentrated) tokens, which effectively suppresses the noise from function words and frequently-occurring nouns while amplifying the signal from semantically discriminative tokens. This directly translates into more precise interval detection, as reflected in the \emph{Entropy (Ours)} row of Table~\ref{tab:ablation} in the main paper.

\begin{figure}[t]
  \centering
  \includegraphics[width=0.85\linewidth]{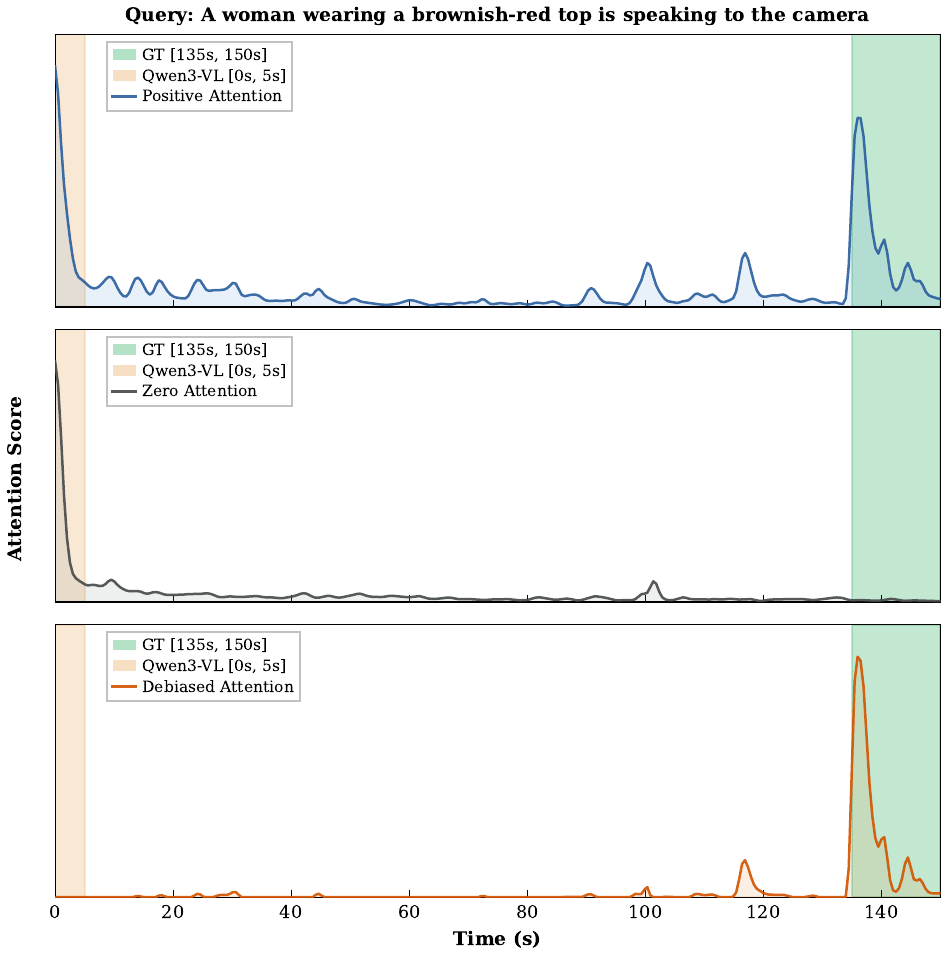}
  \caption{\textbf{Effect of contrastive attention debiasing.} \textbf{Top:} Positive attention on the real video exhibits both a correct peak near the ground-truth interval and a spurious peak at the beginning of the video (attention sink). \textbf{Middle:} Zero attention on a blank video reveals that the beginning-of-video peak is a content-independent bias intrinsic to the model. \textbf{Bottom:} After contrastive debiasing, the bias is removed and only the query-specific peak remains.}
  \label{fig:attention_sink_example}
\end{figure}

\subsection{Contrastive Attention Debiasing}
\label{sec:debiasing_appendix}

Even after entropy-based aggregation, the resulting curve still suffers from systematic content-independent biases. The most prominent one is the well-known ``attention sink'' phenomenon~\citep{kang2025see}: MLLMs tend to allocate a disproportionately large fraction of attention to the first few visual tokens, regardless of what the query is about. Figure~\ref{fig:attention_sink_example} illustrates this effect. The \emph{Positive Attention} curve (top) peaks correctly around the ground-truth interval, but it is accompanied by a tall spurious peak at the very beginning of the video. This spurious peak is purely an artifact of the model's architectural bias: it is still present when the video is replaced by a blank (all-zero) video, as shown by the \emph{Zero Attention} curve (middle). By dividing the positive attention by the zero attention and keeping only the positive contributions, the \emph{Debiased Attention} curve (bottom) removes the attention sink and reveals a clean, query-specific attention signal that can be reliably used for interval detection.

\subsection{Disentangling Temporal Guidance from Resolution Gains}
\label{sec:resolution_control}

For Qwen3-VL and MiMo-VL, cropping under a fixed total visual-token budget
increases the effective number of visual tokens per retained frame.
To test whether our gain mainly comes from this higher resolution, we evaluate
a fixed-token-density variant. Given a crop containing $T'$ of the original
$T$ frames, we reduce the Stage~2 visual budget from $B$ to
\[
B' \approx B \frac{T'}{T},
\]
so that the average number of visual tokens per frame remains approximately
unchanged. The TG-guided interval and all other settings are kept identical.

\begin{table}[ht]
\centering
\caption{
Effect of crop-induced resolution on Qwen3-VL-8B over QVHighlights-TimeLens.
}
\label{tab:resolution_control}
\begin{tabular}{lcccc}
\toprule
Method & R1@0.3 & R1@0.5 & R1@0.7 & mIoU \\
\midrule
Baseline
& 74.1 & 64.1 & 49.1 & 59.4 \\
Crop (Fixed Resolution)
& 77.1 & 66.6 & 51.0 & 61.4 \\
Crop (Ours)
& \textbf{77.3} & \textbf{67.0} & \textbf{51.2} & \textbf{61.9} \\
\bottomrule
\end{tabular}
\end{table}

As shown in Table~\ref{tab:resolution_control}, removing the additional
per-frame resolution still improves mIoU from 59.4 to 61.4, compared with
61.9 for the full method. This indicates that most of the gain comes from
TG-guided temporal selection, while the higher effective resolution provides
a smaller complementary benefit. We note that this resolution effect does not
apply to Molmo2, whose native processor uses a fixed number of visual tokens
per sampled frame.

\section{Extended Evaluation and Reproducibility}

\subsection{Comparison with Other Training-Free Methods}
\label{sec:training_free_comparison}

\begin{table}[ht]
\centering
\caption{\textbf{Comparison with other training-free methods} on QVHighlights-TimeLens. All methods use Qwen3-VL-8B as the backbone. Moment-GPT's captioning models are also replaced with Qwen3-VL-8B for a fair comparison.}
\label{tab:training_free}
\begin{tabular}{lcccc}
\toprule
\textbf{Method} & R1@0.3 & R1@0.5 & R1@0.7 & mIoU \\
\midrule
Baseline          & 74.1 & 64.1 & 49.1 & 59.4 \\
\midrule
Moment-GPT~\citep{xu2025zero}   & 66.5 & 56.1 & 37.9 & 49.8 \\
NumPro~\citep{wu2025number}      & 74.3 & 64.5 & 49.0 & 59.6 \\
Ours                              & \textbf{77.3} & \textbf{67.0} & \textbf{51.2} & \textbf{61.9} \\
\bottomrule
\end{tabular}
\end{table}

To further contextualize our approach, we compare with two representative training-free VTG methods on QVHighlights-TimeLens using Qwen3-VL-8B as the backbone for all methods. We use ``training-free'' here in the sense of requiring no parameter updates. Our method does use a small labeled calibration set once per backbone to identify TG-Heads and set confidence thresholds. Results are reported in Table~\ref{tab:training_free}.

\textbf{Moment-GPT}~\citep{xu2025zero} converts VTG into a text-matching problem: it first generates per-frame captions with an image captioning model, then produces segment-level captions with a video captioning model, and finally scores the similarity between query text and segment captions to select the best temporal span. In our reproduction, we replace both the image captioner and the segment captioner with Qwen3-VL-8B to ensure a fair comparison under the same backbone. As shown in Table~\ref{tab:training_free}, this caption-then-match pipeline performs substantially below the direct decoding baseline ($49.8$ vs.\ $59.4$ mIoU). The bottleneck is that converting continuous visual content into discrete text captions inevitably discards fine-grained temporal information, and errors in caption generation propagate to the matching stage, leading to imprecise boundaries.

\textbf{Number-Prompt (NumPro)}~\citep{wu2025number} overlays a visible frame number onto each video frame and asks the MLLM to ``read'' the temporal position via its OCR capability. This approach was shown to be effective on earlier MLLMs that lacked explicit temporal encoding. However, modern MLLMs such as Qwen3-VL already interleave textual timestamp tokens before each frame's visual tokens, providing the model with an explicit sense of time without relying on visual watermarks. As a result, adding frame-number overlays to Qwen3-VL yields negligible improvement over the baseline ($59.6$ vs.\ $59.4$ mIoU). The timestamp mechanism built into the model architecture already subsumes the information that NumPro attempts to inject visually.

These results suggest that earlier training-free techniques were designed around the limitations of weaker MLLMs (e.g., lack of temporal encoding or poor direct grounding ability). In contrast, our method reads the model's \emph{internal} attention signal rather than manipulating the \emph{external} input (captions or visual watermarks), and achieves a much larger improvement ($+2.5$ mIoU) without modifying the video content or the prompt format.

\subsection{Results on the VTG-Tuned TimeLens Model}
\label{sec:timelens_results}

We additionally apply our framework to TimeLens-8B~\citep{zhang2025timelens}, which is post-trained from Qwen3-VL-8B on 100K temporal-grounding annotations. As shown in Table~\ref{tab:timelens_results}, the method remains beneficial across all three benchmarks, although the gains are smaller than on general-purpose MLLMs because the backbone has already been optimized for timestamp generation.

\begin{table}[ht]
\centering
\caption{\textbf{Results on the VTG-tuned TimeLens-8B model.} Gains are shown in green ($\uparrow$).}
\label{tab:timelens_results}
\setlength{\tabcolsep}{4pt}
\resizebox{\textwidth}{!}{
\begin{tabular}{l|cccc|cccc|cccc}
\toprule
\multicolumn{1}{c}{\multirow{2}{*}[-3pt]{\textbf{Model}}}
& \multicolumn{4}{c}{\textbf{QVHighlights-TimeLens}}
& \multicolumn{4}{c}{\textbf{Charades-TimeLens}}
& \multicolumn{4}{c}{\textbf{ActivityNet-TimeLens}} \\
\cmidrule(lr){2-5}\cmidrule(lr){6-9}\cmidrule(lr){10-13}
& R1@0.3 & R1@0.5 & R1@0.7 & mIoU
& R1@0.3 & R1@0.5 & R1@0.7 & mIoU
& R1@0.3 & R1@0.5 & R1@0.7 & mIoU \\
\midrule
TimeLens-8B & 80.1 & 71.5 & 55.5 & 65.4 & 76.6 & 63.0 & 35.2 & 55.2 & 68.7 & 58.0 & 40.8 & 53.1 \\
TimeLens-8B (+Ours)
& \textbf{80.6}\up{0.5} & \textbf{71.9}\up{0.4} & \textbf{55.7}\up{0.2} & \textbf{65.8}\up{0.4}
& \textbf{76.9}\up{0.3} & \textbf{63.4}\up{0.4} & \textbf{35.3}\up{0.1} & \textbf{55.5}\up{0.3}
& \textbf{68.9}\up{0.2} & \textbf{58.2}\up{0.2} & \textbf{41.2}\up{0.4} & \textbf{53.3}\up{0.2} \\
\bottomrule
\end{tabular}}
\end{table}

\begin{figure}[ht]
  \centering
  \includegraphics[width=1.0\linewidth]{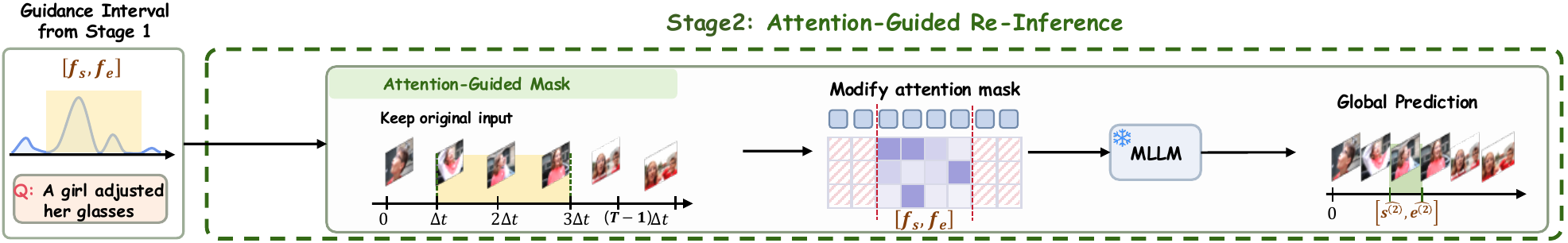}
  \caption{\textbf{Attention-guided Mask for TimeLens-8B.} Instead of cropping and rewriting timestamps, we retain the original input and timeline but prevent all tokens from attending to visual tokens outside the Stage~1 guidance interval $[f_s,f_e]$. The model therefore suppresses distracting frames while producing timestamps directly on the original global timeline.}
  \label{fig:timelens_soft_mask}
\end{figure}

\paragraph{Why Mask for TimeLens-8B?}
Unlike the general-purpose MLLMs evaluated in the main paper, TimeLens-8B is explicitly post-trained for VTG with videos represented on their original timelines and with timestamp tokens interleaved before the visual tokens. The attention-guided crop used in our main framework creates a shorter clip, rewrites its timestamps to start from zero, changes the duration and timestamp-token distribution seen by the model, and removes surrounding temporal context. These changes can disrupt the temporal representation on which a VTG-specialized model has been optimized. We therefore adapt Stage~2 with an attention-guided \emph{Mask}, illustrated in Figure~\ref{fig:timelens_soft_mask}. The original video, timestamp tokens, and input sequence are kept unchanged; we only set the attention-mask entries associated with visual tokens outside $[f_s,f_e]$ to $-\infty$, making those distractor frames invisible to subsequent attention while preserving the original timeline. TimeLens-8B can thus retain its learned timestamp-decoding prior and produce a global prediction directly, without the offset remapping required by cropping.

\subsection{Efficiency Analysis}
\label{sec:efficiency}

Every sample pays one extra zero-video pass for contrastive debiasing, and the confidence gate sends only low-confidence samples to Stage~2. Table~\ref{tab:efficiency} reports per-sample latency measured on one H20 GPU with batch size 1. A skipped sample costs about $2\times$ the baseline, since the zero-video pass repeats the same prefill and decoding a short numeric answer is cheap; a non-skipped sample costs the full three-pass time listed as ``Worst''. Hence $\mathrm{Avg}\approx p\cdot 2\,\mathrm{Base}+(1-p)\cdot\mathrm{Worst}$, where $p$ is the observed skip rate. Because the gate skips $80\%$--$94\%$ of samples, the average overhead is $2.1\times$--$2.4\times$ the baseline rather than the $3.0\times$--$4.8\times$ worst case. Stage~2 costs differ between the two backbones. TimeLens-8B uses Mask, which reuses the original input and changes only the attention mask, adding about one baseline pass. Qwen3-VL-8B uses Crop and adds $2.2\times$--$2.8\times$ a baseline pass. The cropped clip is still encoded to the same token budget $B$, so its prefill cost is the same as a baseline pass; the extra time goes to re-decoding the segment from the video file and processing it at a higher per-frame resolution. This is host-side preprocessing rather than transformer compute, and caching the decoded frames would remove it.

\begin{table}[ht]
\centering
\caption{\textbf{Efficiency analysis.} Per-sample inference time in seconds. \emph{Base} is the standard single-pass baseline, \emph{Worst} the full three-pass pipeline with the gate disabled, and \emph{Avg} the measured average with the gate enabled.}
\label{tab:efficiency}
\setlength{\tabcolsep}{3pt}
\resizebox{\textwidth}{!}{
\begin{tabular}{l|cccc|cccc|cccc}
\toprule
\multicolumn{1}{c}{\multirow{2}{*}[-3pt]{\textbf{Model}}}
& \multicolumn{4}{c}{\textbf{QVHighlights-TimeLens}}
& \multicolumn{4}{c}{\textbf{Charades-TimeLens}}
& \multicolumn{4}{c}{\textbf{ActivityNet-TimeLens}} \\
\cmidrule(lr){2-5}\cmidrule(lr){6-9}\cmidrule(lr){10-13}
& Base & Worst & Avg & Skip\%
& Base & Worst & Avg & Skip\%
& Base & Worst & Avg & Skip\% \\
\midrule
Qwen3-VL-8B & 3.8 & 15.9 & 8.9 & 84\% & 2.8 & 11.7 & 6.6 & 83\% & 3.6 & 17.1 & 8.1 & 91\% \\
TimeLens-8B & 3.8 & 11.6 & 8.4 & 80\% & 2.8 & 8.5 & 5.8 & 94\% & 3.6 & 11.0 & 7.5 & 91\% \\
\bottomrule
\end{tabular}}
\end{table}

\paragraph{Where does the average gain come from?}
The confidence gate deliberately leaves most examples unchanged and routes only low-confidence examples to Stage~2. Consequently, the headline improvement averaged over the entire benchmark understates the effect on the subset where the intervention is actually applied. Let $p_{\mathrm{skip}}$ denote the skip rate. Since skipped examples return exactly the Stage~1 prediction, their per-example IoU is unchanged. Therefore, for mean IoU,
\[
\Delta_{\mathrm{all}}
=
(1-p_{\mathrm{skip}})\Delta_{\mathrm{routed}},
\]
where $\Delta_{\mathrm{routed}}$ is the average improvement on samples routed to Stage~2. For example, on Qwen3-VL-8B/QVHighlights-TimeLens, Stage~2 is invoked for only approximately 16\% of examples, yet the overall benchmark improves by 2.5 mIoU. This corresponds to an average improvement of roughly 15.6 mIoU on the routed subset. Similarly, the modest aggregate gains on the already VTG-tuned TimeLens-8B model arise because 80--94\% of examples are left unchanged. The intervention is therefore concentrated on a relatively small subset of difficult examples rather than uniformly modifying the entire benchmark.

\begin{table}[ht]
\centering
\caption{\textbf{Overall versus routed-subset mIoU gains.} The routed-subset values are implied by the overall gain and skip rate; values are approximate because the displayed aggregate metrics are rounded.}
\label{tab:routed_gain}
\begin{tabular}{llcc}
\toprule
\textbf{Model} & \textbf{Benchmark} &
$\Delta$ \textbf{mIoU (All)} &
$\Delta$ \textbf{mIoU (Routed)} \\
\midrule
TimeLens-8B & QVHighlights & +0.4 & $\approx$ +2.0 \\
TimeLens-8B & Charades     & +0.3 & $\approx$ +5.0 \\
TimeLens-8B & ActivityNet  & +0.2 & $\approx$ +2.2 \\
\midrule
Qwen3-VL-8B & QVHighlights & +2.5 & $\approx$ +15.6 \\
Qwen3-VL-8B & Charades     & +1.3 & $\approx$ +7.6 \\
Qwen3-VL-8B & ActivityNet  & +1.3 & $\approx$ +14.4 \\
\bottomrule
\end{tabular}
\end{table}

\subsection{Hyperparameter Settings}
\label{sec:hyperparams}

Most hyperparameters in Table~\ref{tab:hyperparams} are fixed across models and datasets. Specifically, the number of TG-Heads $K$, entropy temperature $\tau$, Gaussian smoothing $\sigma$, and threshold ratio $\rho$ are shared, as are the frame sampling rate and total visual-token budget for backbones using the TimeLens protocol. Only the dilation window $\delta$ and expansion ratio $r$ vary with dataset duration. Crop is used for the three general-purpose MLLMs, whereas Mask is used for the VTG-tuned TimeLens-8B.

The two confidence thresholds are model-level parameters because both decode confidence and attention confidence are tied to backbone calibration and correlate with localization accuracy (Figure~\ref{fig:analysis}). We use the same 500-example calibration subset described in Appendix~\ref{sec:calibration_stability} for both TG-Head ranking and
confidence-threshold selection. The resulting model-level thresholds are
then applied unchanged to all test sets. A value of $\theta_{\mathrm{attn}} = 1.1$ indicates that the attention-confidence branch is disabled. Because $c_{\mathrm{attn}} \leq 1$ by definition, this threshold can never be triggered, and the gating decision for these backbones depends only on $c_{\mathrm{dec}}$.

\begin{table}[ht]
\centering
\caption{\textbf{Algorithmic hyperparameters} of our framework, with the symbols used in the main paper.}
\label{tab:hyperparams}
\begin{tabular}{lcc}
\toprule
\textbf{Parameter} & \textbf{Symbol} & \textbf{Default} \\
\midrule
Number of TG-Heads & $K$ & 5 \\
Entropy temperature & $\tau$ & 0.5 \\
Gaussian smoothing & $\sigma$ & 2.0 \\
Threshold ratio & $\rho$ & 0.2 \\
Dilation window (seconds) & $\delta$ & 5.0 \\
Expansion ratio & $r$ & 0.15 \\
Decode confidence threshold & $\theta_{\text{dec}}$ & 0.9 \\
Attention confidence threshold & $\theta_{\text{attn}}$ & 0.9 \\
\bottomrule
\end{tabular}
\end{table}

\begin{table}[ht]
\centering
\caption{\textbf{Final hyperparameter configuration for all evaluated backbones.} FPS is the frame sampling rate, ``Visual budget'' is the total visual-token budget for MiMo-VL/Qwen3-VL/TimeLens and the maximum number of frames for Molmo2, and ``Stage 2'' is the chosen Stage~2 variant. A value of $\theta_{\mathrm{attn}} = 1.1$ indicates that the attention-confidence branch is disabled. Confidence thresholds are selected using the same 500-example calibration subset drawn from the TimeLens-100K training corpus.}
\label{tab:model_dataset_hyperparams}
\setlength{\tabcolsep}{4pt}
\resizebox{\textwidth}{!}{
\begin{tabular}{ll|cc|c|cccccccc}
\toprule
\textbf{Model} & \textbf{Dataset} & \textbf{FPS} & \textbf{Visual budget} & \textbf{Stage~2} & $K$ & $\tau$ & $\sigma$ & $\rho$ & $\delta$(s) & $r$ & $\theta_{\text{dec}}$ & $\theta_{\text{attn}}$ \\
\midrule
\multirow{3}{*}{MiMo-VL-7B}
& QVHighlights-TimeLens & 2 & 14{,}336 & Crop & 5 & 0.5 & 2.0 & 0.2 & 5.0 & 0.15 & 0.9 & 0.9 \\
& Charades-TimeLens     & 2 & 14{,}336 & Crop & 5 & 0.5 & 2.0 & 0.2 & 3.0 & 0.1 & 0.9 & 0.9 \\
& ActivityNet-TimeLens  & 2 & 14{,}336 & Crop & 5 & 0.5 & 2.0 & 0.2 & 5.0 & 0.15 & 0.9 & 0.9 \\
\midrule
\multirow{3}{*}{Qwen3-VL-8B}
& QVHighlights-TimeLens & 2 & 14{,}336 & Crop & 5 & 0.5 & 2.0 & 0.2 & 5.0 & 0.15 & 0.8 & 0.9 \\
& Charades-TimeLens     & 2 & 14{,}336 & Crop & 5 & 0.5 & 2.0 & 0.2 & 3.0 & 0.1 & 0.8 & 0.9 \\
& ActivityNet-TimeLens  & 2 & 14{,}336 & Crop & 5 & 0.5 & 2.0 & 0.2 & 5.0 & 0.15 & 0.8 & 0.9 \\
\midrule
\multirow{3}{*}{Molmo2-8B}
& QVHighlights-TimeLens & 2 & 384 frames & Crop & 5 & 0.5 & 2.0 & 0.2 & 6.0 & 0.3 & 0.7 & 1.1 \\
& Charades-TimeLens     & 2 & 384 frames & Crop & 5 & 0.5 & 2.0 & 0.2 & 4.0 & 0.15 & 0.7 & 1.1 \\
& ActivityNet-TimeLens  & 2 & 384 frames & Crop & 5 & 0.5 & 2.0 & 0.2 & 4.0 & 0.05 & 0.7 & 1.1 \\
\midrule
\multirow{3}{*}{TimeLens-8B}
& QVHighlights-TimeLens & 2 & 14{,}336 & Mask & 5 & 0.5 & 2.0 & 0.2 & 5.0 & 0.15 & 0.8 & 1.1 \\
& Charades-TimeLens     & 2 & 14{,}336 & Mask & 5 & 0.5 & 2.0 & 0.2 & 3.0 & 0.1 & 0.8 & 1.1 \\
& ActivityNet-TimeLens  & 2 & 14{,}336 & Mask & 5 & 0.5 & 2.0 & 0.2 & 5.0 & 0.15 & 0.8 & 1.1 \\
\bottomrule
\end{tabular}}
\end{table}

\subsection{Full Implementation Details}
\label{sec:impl_details_appendix}

Throughout this work we aim for a direct comparison with TimeLens~\citep{zhang2025timelens}, so we keep the inference protocol identical to theirs whenever possible. We summarize the full details below for reproducibility.

\paragraph{Prompt template.}
We use the same grounding prompt as TimeLens~\citep{zhang2025timelens}. The entire conversation consists of a single user turn that contains the video followed by an instruction:

\begin{quote}\small
\textbf{User:} \textless video\textgreater\\
Please find the visual event described by the sentence `\{query\}', determining its starting and ending times. The format should be: `The event happens in \textless start time\textgreater\ -\ \textless end time\textgreater\ seconds'.
\end{quote}

Qwen3-VL-8B and TimeLens-8B use Qwen3-VL's native interleaved timestamp mechanism. Molmo2-8B likewise uses its native processor, which inserts a textual wall-clock timestamp before each frame's visual tokens. For MiMo-VL-7B, we additionally append the \texttt{/no\_think} suffix to disable its reasoning mode, following its official inference recipe.

\paragraph{Video sampling and token budget.}
Each video is sampled at 2 FPS and passed to the backbone's native video processor. For Qwen3-VL-based backbones (Qwen3-VL and TimeLens-8B), the encoder uses a spatial patch downsample rate of 32; for MiMo-VL-7B and Qwen2.5-VL-based backbones, the rate is 28. These models use a total visual-token budget of $B=14{,}336$, matching the configuration adopted by TimeLens. Molmo2-8B instead follows its native \texttt{uniform\_last\_frame} sampling, with at most 384 frames at $378\times378$ resolution and 81 visual tokens per frame. Per-model settings are listed in Table~\ref{tab:model_dataset_hyperparams}.

\paragraph{Decoding.}
All predictions (the Stage 1 baseline, the Stage 2 re-inference, and all ablations) are produced with greedy decoding, temperature $0$, and at most $64$ new tokens. We extract the two floating-point timestamps from the decoded string with a simple regular-expression matcher, following the same post-processing used by TimeLens.

\paragraph{Hardware and software.}
The experiments are conducted on a node with 8$\times$H20 GPUs. Each model instance runs on one GPU with batch size 1, and each test set is sharded across eight GPUs for parallel evaluation. We use PyTorch with SDPA attention kernels; our attention-extraction and attention-mask implementations are written as light-weight monkey patches over the SDPA call so that they incur no additional GPU memory beyond the baseline forward pass.

\paragraph{TG-Head identification.}
For each backbone, we run the attention-knockout procedure of Section~\ref{sec:tghead} once on a held-out subset of 500 samples uniformly drawn from TimeLens's 100K training corpus~\citep{zhang2025timelens}. The resulting top-5 GCS heads are fixed and reused across all three benchmarks. Their layer--head indices are reported in Appendix~\ref{sec:tg_heads_models}. The 500 calibration samples come from the TimeLens-100K training corpus, whose videos originate from sources such as CosMoCap, InternVid-VTime, DiDeMo, QuerYD, and HiREST. These sources are disjoint from the three evaluation benchmarks used in this paper: Charades-TimeLens (built on Charades-STA), ActivityNet-TimeLens (built on ActivityNet Captions), and QVHighlights-TimeLens (built on QVHighlights). We use 500 samples as a conservative default; Appendix~\ref{sec:calibration_stability} shows that reducing the calibration set to 200 examples selects the identical top-5 heads and yields unchanged downstream performance.


\section{Qualitative Analysis}

\subsection{More Successful Examples}
\label{sec:more_vis}

\begin{figure}[ht]
  \centering
  \includegraphics[width=0.8\linewidth]{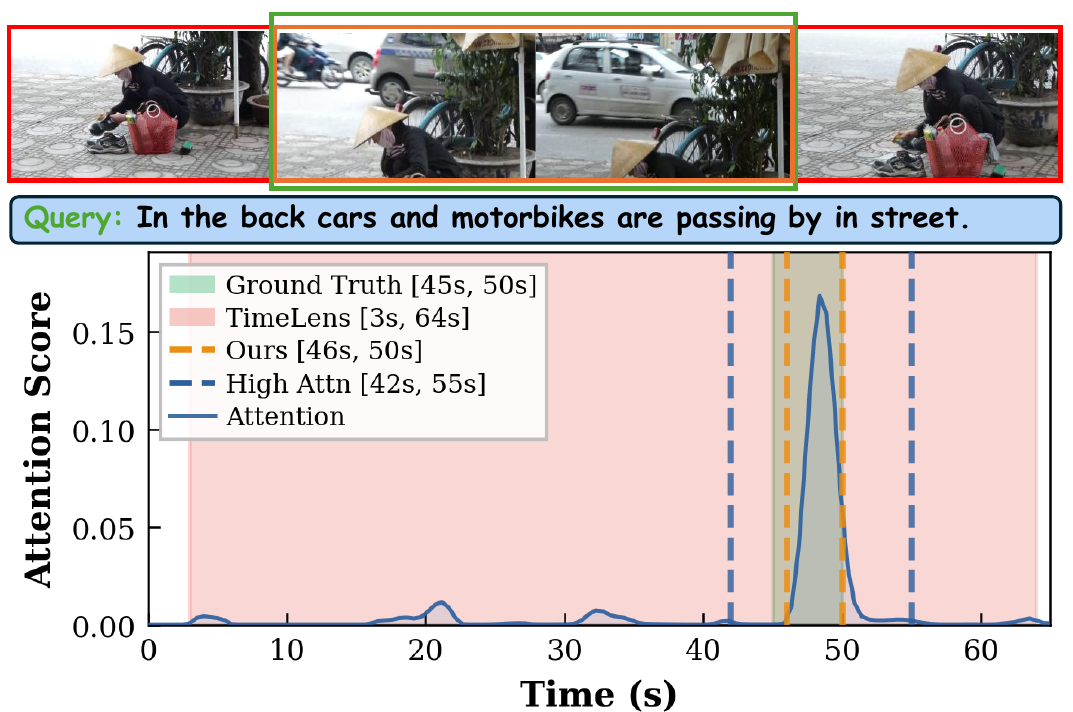}
  \caption{\textbf{Successful correction (Example 1).} The baseline predicts an overly broad interval that covers almost the entire video, while the TG-Head attention concentrates on a narrow high-attention region. Restricting the visual context to this region allows Stage 2 to localize the event much more precisely.}
  \label{fig:more_vis_1}
\end{figure}

\begin{figure}[ht]
  \centering
  \includegraphics[width=0.8\linewidth]{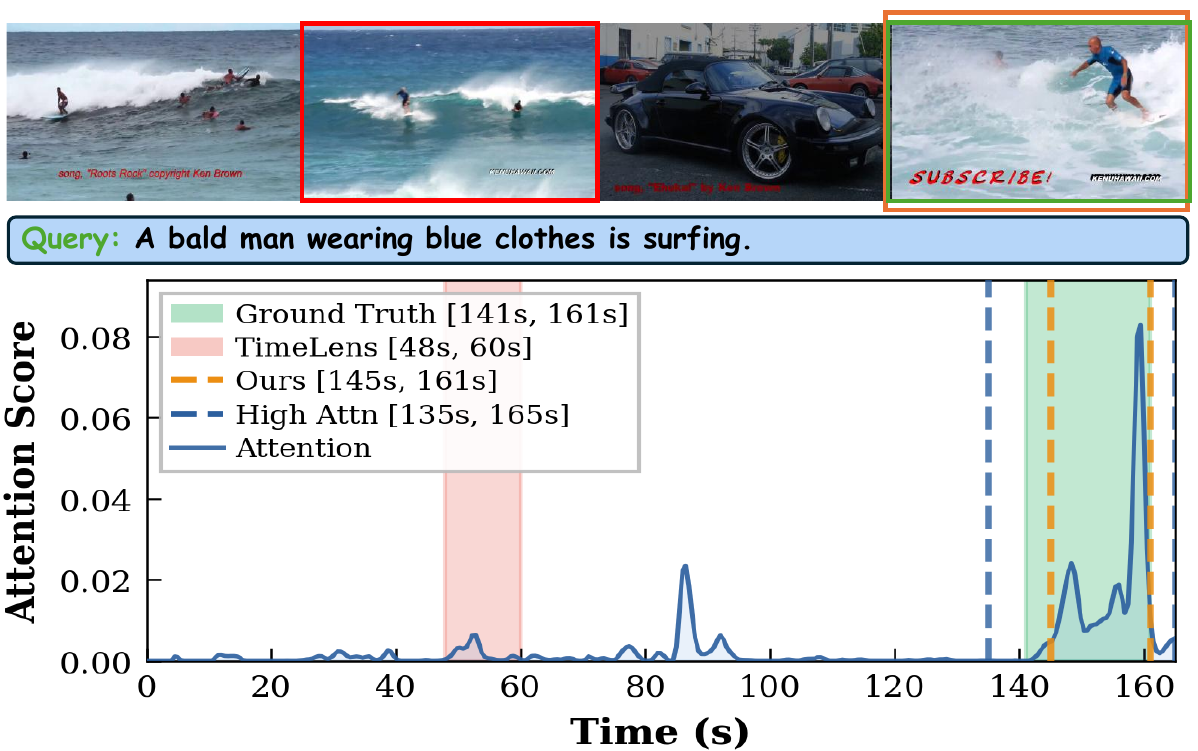}
  \caption{\textbf{Successful correction (Example 2).} The baseline produces an inaccurate interval, while our method tightens it around the true event by leveraging the clean prefill attention signal.}
  \label{fig:more_vis_2}
\end{figure}

We provide additional qualitative examples (Figures~\ref{fig:more_vis_1},~\ref{fig:more_vis_2}) beyond those in the main paper. Example~1 shows that our method can tighten an overly broad baseline prediction by focusing on the narrow high-attention region. Example~2 shows that TG-Head attention focuses on the true surfing segment, allowing re-inference to refine the inaccurate baseline prediction.

\subsection{Failure Cases}
\label{sec:failure_cases}

While our framework consistently improves average performance, there are individual samples on which it fails to help or even slightly hurts the prediction. We present two representative failure modes below.

\begin{figure}[ht]
  \centering
  \includegraphics[width=1.0\linewidth]{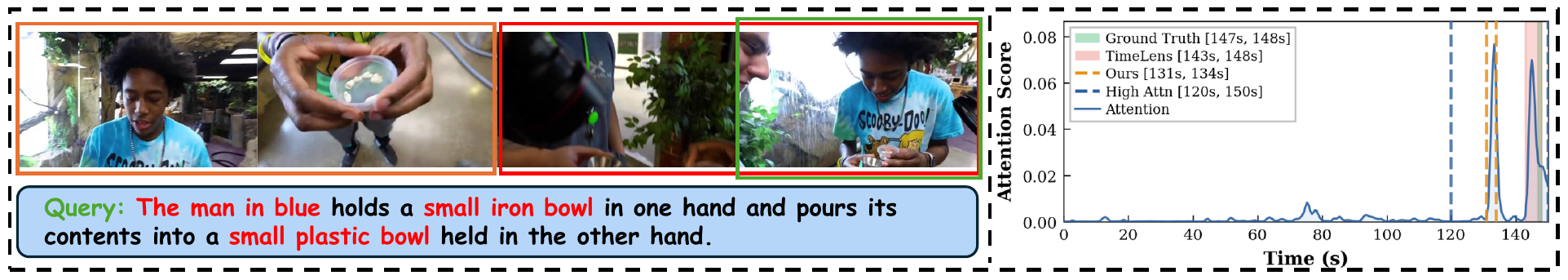}
  \caption{\textbf{Failure case 1 (multi-peak).} The query mentions a ``small plastic bowl'' that appears at two separate moments. Stage 1 correctly picks the later peak (ground truth), but the detected high-attention interval encompasses both peaks and Stage 2 is pulled to the earlier, higher-amplitude one.}
  \label{fig:badcase}
\end{figure}

The first mode (Figure~\ref{fig:badcase}) is the multi-peak case: a queried object or action recurs at several moments in the video, producing multiple attention peaks. The high-attention interval merges these peaks, and Stage 2 can be pulled toward the wrong one even though Stage 1 happened to pick the correct peak.

\begin{figure}[ht]
  \centering
  \includegraphics[width=0.8\linewidth]{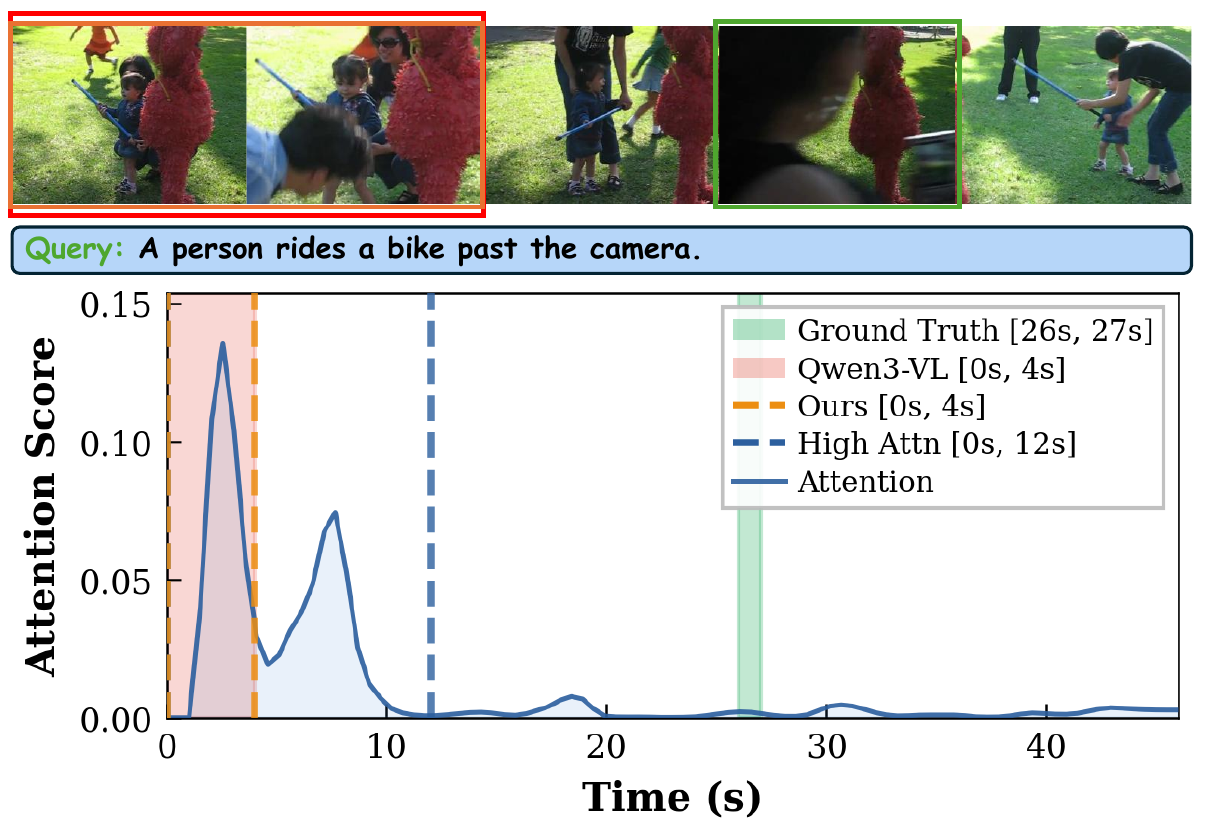}
  \caption{\textbf{Failure case 2 (absent attention at GT).} The ground-truth interval receives near-zero attention, while both the baseline and our method predict a high-attention but incorrect segment. When TG-Head attention does not correlate with the ground truth, our framework cannot provide a correction.}
  \label{fig:more_vis_fail}
\end{figure}

The second mode (Figure~\ref{fig:more_vis_fail}) is more fundamental: the ground-truth interval itself receives little or no attention from the TG-Heads. In this case, the attention curve peaks at a different segment that is visually salient but does not correspond to the queried event. Both the baseline model and our method are drawn to this high-attention but incorrect interval, because our framework fundamentally relies on the assumption that TG-Head attention correlates with the ground truth. For the small fraction of samples where this assumption fails, our method cannot provide a correction.

\section{Limitations and Future Work}
\label{sec:limitations}

\paragraph{Limitations.}
Our framework has two main limitations. First, the high-attention interval detection assumes the query corresponds to a single contiguous temporal region. This matches the current standard VTG benchmarks, which are all single-interval, but may not generalize to multi-interval settings. Even within single-interval benchmarks, when a key object or action recurs at several moments in the video (see Appendix~\ref{sec:failure_cases} for concrete examples), our detector may merge the peaks and Stage 2 may be pulled toward the wrong one. Second, the framework adds extra forward passes on top of baseline inference: one zero-video pass for bias estimation, and one Stage 2 pass for samples that the confidence gate does not skip. With the confidence gate enabled, this increases average per-sample inference time by approximately $2.1\times$--$2.4\times$ (see Appendix~\ref{sec:efficiency}), which is acceptable for offline evaluation but may be non-trivial for latency-critical applications.

\paragraph{Future work.}
Three directions seem particularly promising. (i) \emph{Multi-interval detection.} Extending our detector to return multiple candidate intervals and letting Stage 2 evaluate each of them would address the multi-peak failure mode discussed above. (ii) \emph{Beyond temporal grounding.} The perception-generation gap we observe is not specific to VTG; similar patterns have been reported on visual question answering~\citep{liu2025seeing, zhang2025mllms}. We plan to explore whether the same ``read-then-regenerate'' recipe can improve other tasks such as dense video captioning, temporal action detection, or even spatial grounding in images. (iii) \emph{Merging Stage 2 with post-training.} It would be interesting to study whether explicitly supervising a post-trained model with the attention signal that Stage 1 extracts (rather than only with the ground-truth timestamps) can give the model more concentrated TG-Heads and thus make it better suited to our framework.

\end{document}